%% file: diffusion_fusion_plus.tex
\documentclass{article}
\usepackage{amssymb}
\usepackage{amsmath}
\usepackage{graphicx}
\usepackage{wrapfig}
\usepackage{xcolor}
\usepackage{xspace}
\usepackage[pagebackref]{hyperref}
\usepackage{subcaption}
\usepackage{multirow}
\usepackage{threeparttable}
\usepackage[final]{corl_2025} %
\usepackage{caption}

\usepackage{todonotes}
\usepackage{booktabs}
\usepackage[capitalize]{cleveref}
\crefname{section}{Sec.}{Secs.}
\Crefname{section}{Section}{Sections}
\Crefname{table}{Table}{Tables}
\crefname{table}{Tab.}{Tabs.}

\input{notation}
\usepackage{collcell}
\usepackage{makecell}
\usepackage{colortbl}
\usepackage{tabularx}
\usepackage{colortbl}
\usepackage{pgf}
\usepackage{arydshln}
\definecolor{gnbu}{RGB}{123,204,196}

\DeclareRobustCommand\onedot{\futurelet\@let@token\@onedot}
\def\onedot{\ifx\@let@token.\else\null\fi\xspace}

\def\ie{\emph{i.e.}\onedot}

\def\wrt{w.r.t\onedot}

\def\method{\emph{UnPose}\onedot}

\title{\method: \emph{\underline{Un}}certainty-Guided Diffusion Priors for Zero-Shot \emph{\underline{Pose}} Estimation}

\author{
  Zhaodong Jiang\textsuperscript{1,2}\thanks{Work done during an internship at Huawei Noah's Ark Lab} \enspace
  Ashish Sinha\textsuperscript{1} \enspace
  Tongtong Cao\textsuperscript{1} \enspace
  Yuan Ren\textsuperscript{1} \enspace
  Bingbing Liu\textsuperscript{1} \enspace
  Binbin Xu\textsuperscript{1}\thanks{Corresponding author}\\
  \textsuperscript{1}Huawei Noah's Ark Lab, Canada \qquad
  \textsuperscript{2}University of Toronto, Canada \\
  \texttt{\{firstname.lastname\}@huawei.com} \quad
  \texttt{zhaodong.jiang@mail.utoronto.ca} 
}

\begin{document}
\maketitle

\begin{figure}[h]
    \centering
    \includegraphics[width=0.95\linewidth]{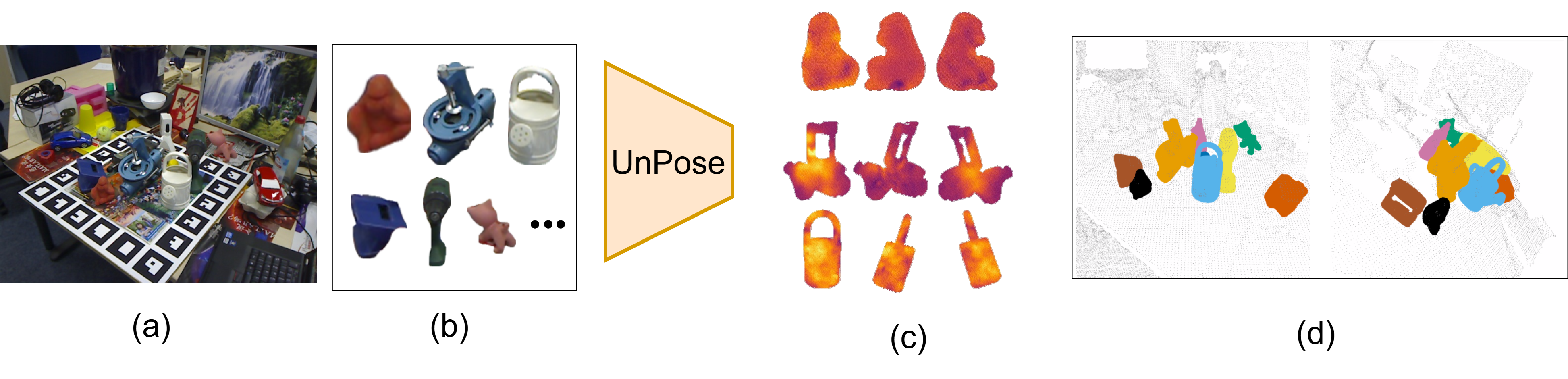}
    \caption{We present~\method for zero-shot model-free pose estimation, that takes (a) RGB-D frames and (b) object segments in the scene, to output (c) 3D priors and uncertainty from a diffusion model to (d) reconstruct and continually refine the geometry of the object represented as 3DGS~\cite{3dgs} for 6D pose estimation. (d) shows the resulting object poses with globally consistent front and back views from the input viewpoint. }
    \label{fig:teaser}
\end{figure}

\begin{abstract}
    Estimating the 6D pose of novel objects is a fundamental yet challenging problem in robotics, often relying on access to object CAD models. 
    However, acquiring such models can be costly and impractical. 
    Recent approaches aim to bypass this requirement by leveraging strong priors from foundation models to reconstruct objects from single or multi-view images, but typically require additional training or produce hallucinated geometry.
    To this end, we propose \method, a novel framework for zero-shot, model-free 6D object pose estimation and reconstruction that exploits 3D priors and uncertainty estimates from a pre-trained diffusion model. 
    Specifically, starting from a single-view RGB-D frame, \method uses a multi-view diffusion model to estimate an initial 3D model using 3D Gaussian Splatting (3DGS) representation, along with pixel-wise epistemic uncertainty estimates.
    As additional observations become available, we incrementally refine the 3DGS model by fusing new views guided by the diffusion model’s uncertainty,
    thereby, continuously improving the pose estimation accuracy and 3D reconstruction quality. 
    To ensure global consistency, the diffusion prior-generated views and subsequent observations are further integrated in a pose graph and jointly optimized into a coherent 3DGS field.
    Extensive experiments demonstrate that \method significantly outperforms existing approaches in both 6D pose estimation accuracy and 3D reconstruction quality.
    We further showcase its practical applicability in real-world robotic manipulation tasks. Video demos can be found at our project page:~\url{https://frankzhaodong.github.io/UnPose}.

\end{abstract}

\keywords{6D Pose Estimation, Diffusion Model, Object Reconstruction, Uncertainty Estimation}

\section{Introduction}

\input{intro}

\begin{figure}[htbp]
    \centering
    \includegraphics[width=0.9\textwidth]{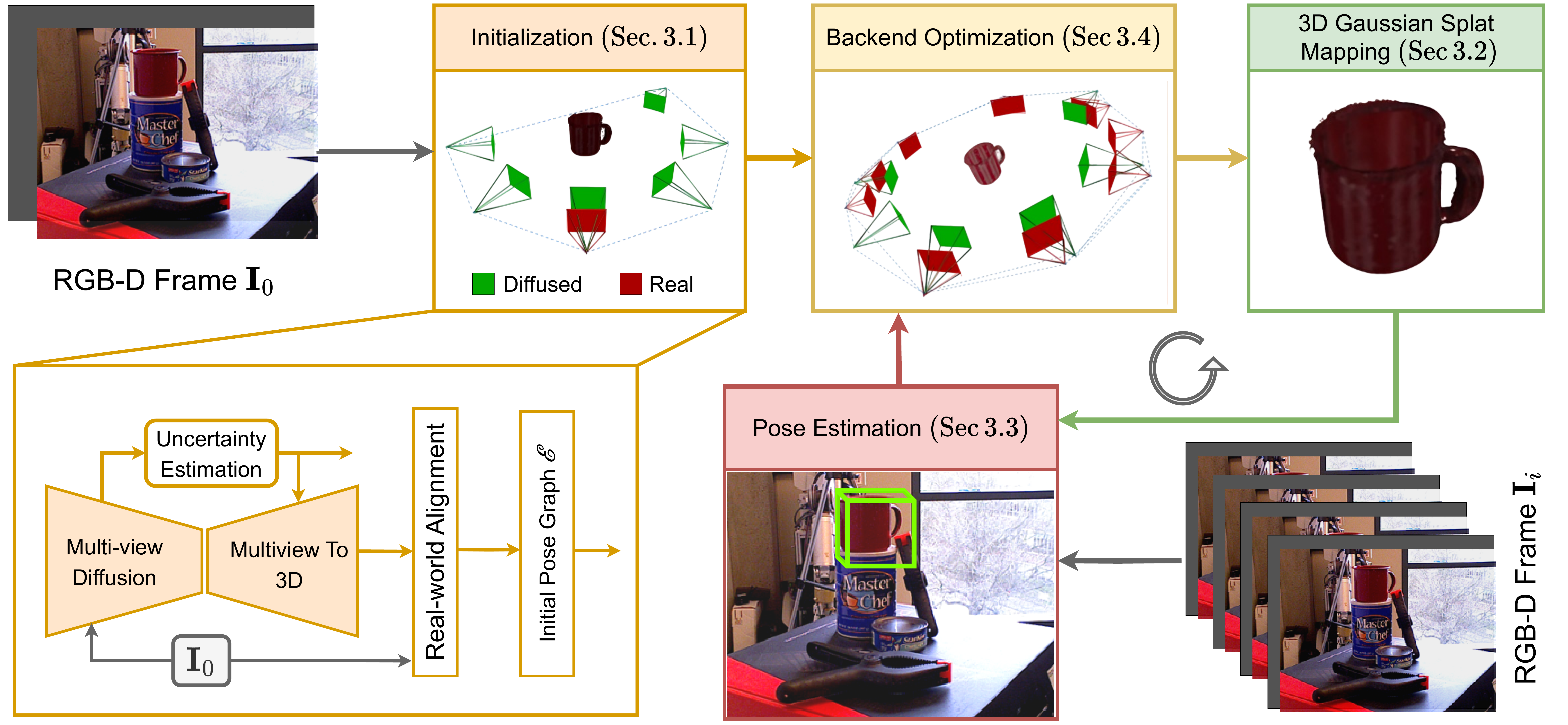}
    \caption{We present an overview of our proposed pipeline. Starting from a single-view RGB-D frame, \method uses a multi-view diffusion model to estimate an initial 3D model using 3D Gaussian Splatting (3DGS) representation, along with pixel-wise epistemic uncertainty estimates.
    As additional observations become available, we incrementally refine the 3DGS model by fusing new views guided by the diffusion model’s uncertainty,
    thereby, continuously improving the pose estimation accuracy and 3D reconstruction quality.
    }
    \label{fig:overview}
\end{figure}

\section{Related Works}
\label{sec:relatedworks}

\subsection{Model-based Object Pose Estimation}
\label{sec:cad_rel}
Instance-level pose estimation techniques~\cite{he2020pvn3d,park2019pix2pose,labbe2020cosypose,he2021ffb6d} typically rely on textured CAD models for specific objects, where both training and testing are conducted on the same set of instances. 
In contrast, category-level methods~\cite{zhang2024omni6dpose,chen2020learning,wang2019normalized,lin2024instance,chen2024secondpose} aim to generalize to \emph{unseen} instances within \emph{known} categories, though they require costly category-specific annotations and are limited to predefined object classes.
To address these limitations, category-agnostic approaches~\cite{caraffa2024freeze,sam6d,chen2024zeropose} aim to estimate the poses of novel objects without relying on predefined categories.
However, many of these methods still depend on ground-truth CAD models to render reference views during inference, which restricts their practical use in open-world applications where such models are often unavailable or costly to obtain.
More recent works~\cite{cai2024gs,ponimatkin20256d} attempt to address this gap by retrieving CAD models from large repositories, though retrieval accuracy and high quality datasets continue to pose challenges.

\subsection{Model-free Object Pose Estimation}
\label{sec:modelfree_rel}
To reduce reliance on textured CAD models, recent research has shifted toward \textit{model-free} object pose estimation methods using either reference images~\cite{liu2024unopose,sam6d,lee2025any6d} 
or single/multi-view images or videos~\cite{bundlesdf,ponimatkin20256d}. 
However, current SOTA approaches still rely on posed reference views \cite{sam6d} or reconstructed object models \cite{fp} and, thus, are not \textit{strictly model free}.
Several works attempt to relax this requirement by estimating relative orientation from a single anchor image~\cite{nguyen2024nope,lee2025any6d}, using feature-matching~\cite{sun2021loftr} or language guidance~\cite{corsetti2024open}.
Despite these advances, performance degrades
when query views have limited overlap or severe occlusions.
To tackle occlusion, recent works~\cite{gigapose,lee2025any6d,liu2025hippoharnessingimageto3dpriors} leverage image-to-3D diffusion priors to reconstruct object's geometry from limited views and estimate the 6D pose, alleviating the need for object-specific CAD models. 
Yet, these methods do not refine the generated 3D models to correct diffusion-induced hallucinations~\cite{aithal2024understanding}, often producing inaccurate geometry and appearance for unobserved regions.
Another direction draws on object-level SLAM~\cite{mccormac2018fusion++,xu2019mid,nicholson2018quadricslam}, which incrementally refines object's geometry over time and handles occlusions.
For example, GOM~\cite{gom} combines diffusion priors with multi-view sensor data in an alternative optimization scheme. %
However, the alternative optimizations between diffusion and NeRF~\cite{nerf} yields suboptimal reconstruction quality.
To address these challenges, we propose a \textit{model-free} approach that combines diffusion priors and observations into an incrementally refined 3DGS~\cite{3dgs} representation.
We incorporate them into a factor graph~\cite{dellaert2017factor} to ensure a globally consistent object representation that leads to robust and accurate object pose estimations.

\subsection{Uncertainty Estimation}
\label{sec:rel_uncertain}

Uncertainty has been widely studied in robotics and computer vision communities~\cite{Thrun:etal:Book2005, kendall2017uncertainties} to improve the robustness and accuracy under noisy (\textit{aleatoric uncertainty}) or incomplete (\textit{epistemic uncertainty}) observations~\cite{liao2024uncertainty,tan2025uncertainty}.
We focus on epistemic uncertainty due to occlusions, ambiguities and limited viewpoints.
A common estimation approach is Deep Ensembles, which captures uncertainty via prediction variance across independently initialized models~\cite{lakshminarayanan2017simple}.
Alternatively, Bayesian Neural Networks~\cite{jospin2022hands} model a posterior distribution over weights, typically using variational inference with KL regularization to approximate the true posterior~\cite{blundell2015weight}.
While effective, these methods are costly, which recent works circumvent through post-hoc Laplace Approximations (LA)~\cite{ritter2018scalable}.
However, applying Bayesian uncertainty to image-to-3D diffusion models (DMs) is challenging due to their high-dimensional parameter space and iterative reverse process, which complicates uncertainty propagation. 
Additionally, DMs are known to hallucinate~\cite{aithal2024understanding}, potentially compromising modeling accuracy. 
To address this, we model epistemic uncertainty in image-to-3D DMs using efficient Bayesian inference~\cite{kou2023bayesdiff}, and further refined through future observations.

\section{Method}
\label{sec:method}

Our proposed pipeline, visualized in~\cref{fig:overview}, consists of four modules: initialization (\cref{sec:init}), pose estimation (\cref{sec:pose}), Backend Optimization (\cref{sec:obs_update}), 3DGS mapping (\cref{sec:gau_map}). 
For each object $\mathcal{O}$ in a scene, we continuously refine its 3D model, represented as a 3DGS field $\boldsymbol{Q}_{o}$~\cite{3dgs}, in its own object canonical coordinate frame $\cframe{O}$, along with a relative pose, $\T{O}{C}$, from the camera coordinate frame, $\cframe{C}$, to the object coordinate frame $\cframe{O}$. Camera poses, $\T{W}{C} \in \mathrm{SE}(3)$, are known using an off-the-shelf SLAM method~\cite{campos2021orb}.

\subsection{Initialization}
\label{sec:init}
Without relying on any CAD models or pre-reconstructed models, our system initializes from an arbitrary RGB-D frame $\mbc{I}_{0} = \{ \mbf{I}_{0} \in \mathbb{R}^{w \times h \times 3}, \mbf{D}_{0} \in \mathbb{R}^{w \times h }, \mbf{M} \in \{0,1\}^{w \times h \times m}\}$ 
by synthesizing multiview diffusion images $\mbfhat{I}_{1:k}$ using the Wonder3D~\cite{wonder3d} where $k$ is set to be 6.
We assume that object masks $\mbf{M}$ are provided to us beforehand for $m$ objects in the scene using a SOTA instance segmentation model.
We further estimate the corresponding pixel-wise uncertainty $\operatorname{Var}(\mbfhat{I})_{1:k} \in \mathbb{R}^{w \times h }$ for the diffused views and align them together with the initial frame $\mbc{I}_{0}$ following VGGT~\cite{wang2025vggt}.
The aligned relative poses $\T{C_0}{C_k}$ together with the estimated uncertainties are refined using pose graph optimization to reconstruct an initial 3DGS object field $\boldsymbol{Q}_{o}$~\cite{3dgs}. %

\begin{figure}
    \centering
    \begin{minipage}{0.43\textwidth}
        \centering
            \includegraphics[width=\linewidth]{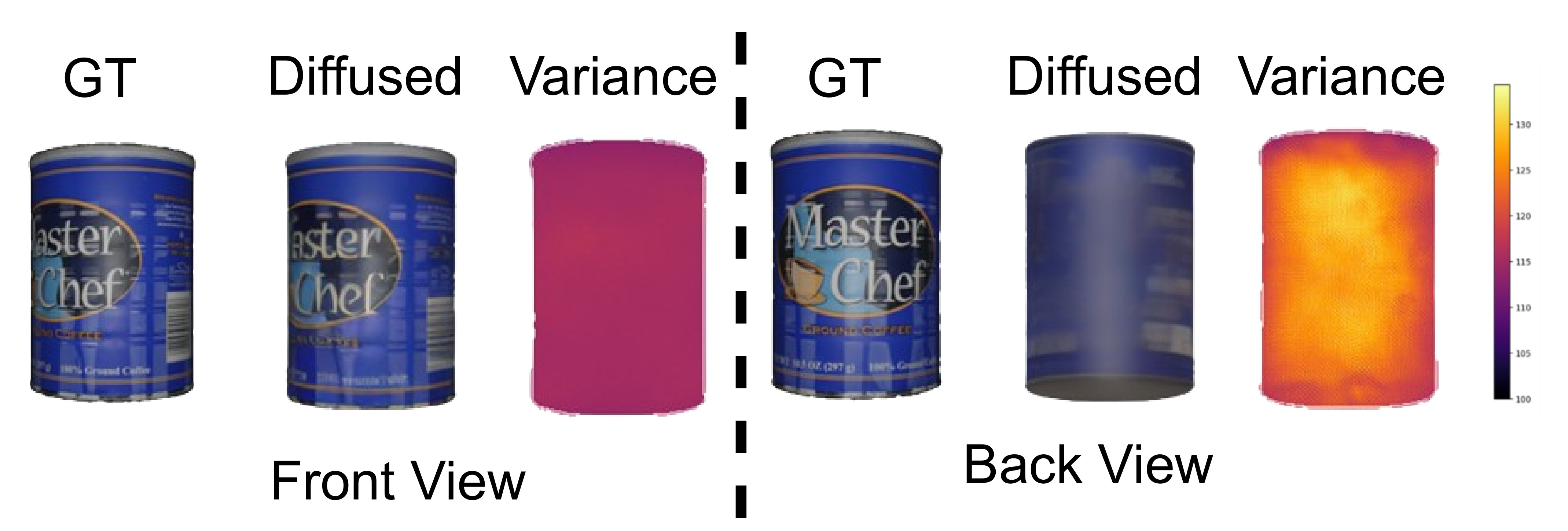}
            \caption{Visualization of ground-truth rendering, and its corresponding diffused view and pixel-level uncertainties extracted from Wonder3D~\cite{wonder3d} using LLLA~\cite{kristiadi2020being}. The diffused images show larger variance at unseen perspective.}
            \label{fig:uncertainity}
    \end{minipage}
    \hfill
    \begin{minipage}{0.55\textwidth}
        \centering
        \includegraphics[width=\linewidth]{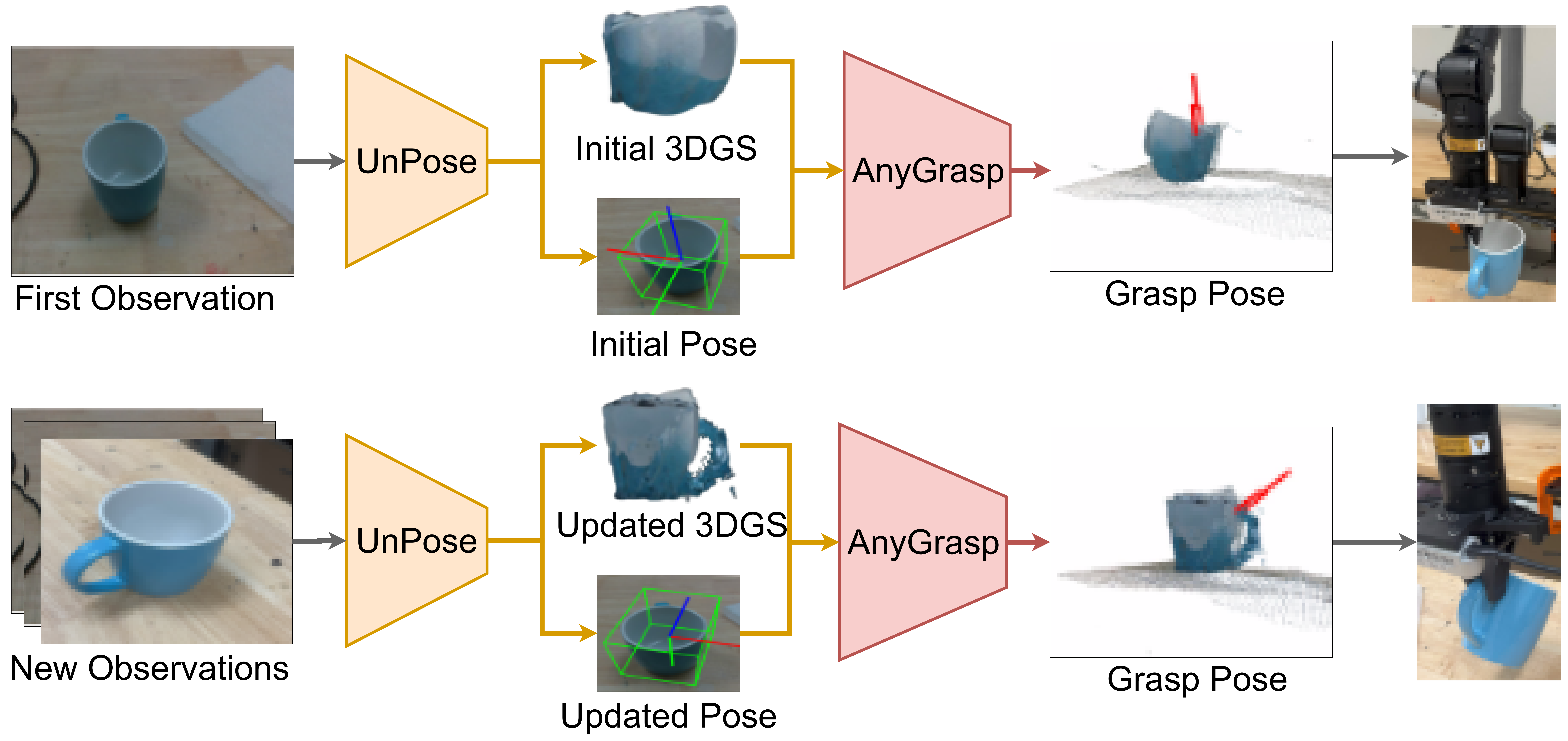}
        \captionof{figure}{Illustration of~\method for a real-world robotic manipulation task.}
        \label{fig:robot}
    \end{minipage}
\end{figure}

\paragraph{Uncertainty Estimation.}
To estimate uncertainty from a pre-trained diffusion model (DM) without additional training, we adapt the approach proposed in BayesDiff~\cite{kou2023bayesdiff} into the Wonder3D framework~\cite{wonder3d}.
Specifically, we adopt last-layer Laplace approximation (LLLA)~\cite{kristiadi2020being} to approximate the predictive distribution of the DM's noise prediction for efficient Bayesian inference, enabling the DM to produce pixel-wise uncertainty alongside synthesized multi-view images. %

Following LLLA~\cite{kristiadi2020being}, we approximate the distribution of the noise prediction $\epsilon_t$  of a pre-trained DM given the noisy state $\mathbf{x}_t$ at diffusion timestep $t$ as a Gaussian distribution:
\begin{equation}
    p(\epsilon_t \mid \mathbf{x}_t, t) \approx \mathcal{N}\left(\epsilon_\theta(\mathbf{x}_t, t), \mathbf{\Sigma}{\epsilon_t} \right),
\end{equation}
where $\theta$ represents the parameters of DM, $\epsilon_\theta(\mathbf{x}_t, t)$ is the standard prediction of the pre-trained model, and $\mathbf{\Sigma}{\epsilon_t}$ is the predictive covariance derived from the LLLA appplied to the model's last layer parameters (see~\cite{kou2023bayesdiff, kristiadi2020being} for more details). For pixel-wise uncertainty, we are primarily interested in the diagonal elements of the covariance, $\operatorname{Var}\left( \epsilon_t \right) = \operatorname{diag}\left(\mathbf{\Sigma}{\epsilon_t}\right)$, visualized in~\cref{fig:uncertainity}.

Wonder3D~\cite{wonder3d} employs DDIM~\cite{song2020denoising} for its deterministic sampling process. The standard DDIM update step to obtain $\mathbf{x}_{t-1}$ from $\mathbf{x}_{t}$ is given by 
\begin{equation}
    \mathbf{x}_{t-1} = \sqrt{\alpha_{t-1}} \left( \frac{\mathbf{x}_t - \sqrt{1 - \alpha_t} \, \epsilon_t(\mathbf{x}_t)}{\sqrt{\alpha_t}} \right)
+ \sqrt{1 - \alpha_{t-1} - \sigma_t^2} \cdot \epsilon_t(\mathbf{x}_t),
\label{eq:ddim_step}
\end{equation}
where $\alpha_{t}$ and $\sigma_{t}$ are noise schedule parameters.

To propagate uncertainty through the denoising steps, we iteratively update the variance $\operatorname{Var}(\mathbf{x}_{t-1})$ based on the variance from the previous step $\operatorname{Var}(\mathbf{x}_{t})$ and the uncertainty from the noise prediction $\operatorname{Var}\left(\epsilon_t\right)$. The variance update corresponding to the DDIM step, \cref{eq:ddim_step}, is:
\begin{align}
    \operatorname{Var}(\mathbf{x}_{t-1}) = &\frac{\alpha_{t-1}}{\alpha_t} \operatorname{Var}(\mathbf{x}_t) 
- 2\frac{\sqrt{\alpha_{t-1}}}{\sqrt{\alpha_{t}}} (\sqrt{1-\alpha_{t-1}-\sigma_t^2}-\frac{\sqrt{\alpha_{t-1}}}{\alpha_t}\sqrt{1-\alpha_t}) \operatorname{Cov}\left(\mathbf{x}_t, \epsilon_t\right) \nonumber \\
&+ (\sqrt{1-\alpha_{t-1}-\sigma_t^2}-\frac{\sqrt{\alpha_{t-1}}}{\alpha_t}\sqrt{1-\alpha_t})^2 \operatorname{Var}\left(\epsilon_t \right) 
\end{align}

The key challenge is estimating the covariance term $\operatorname{Cov}\left(\mathbf{x}_t, \epsilon_t\right)$, which captures the correlation between the noisy state and the model's prediction.
As this is analytically intractable, we estimate it using Monte Carlo sampling~\cite{hastings1970monte}:
\begin{equation}
    \operatorname{Cov}\left(\mathbf{x}_t, \epsilon_\theta\left(\mathbf{x}_t, t \right)\right) 
\approx \frac{1}{S} \sum_{i=1}^S \left( \mathbf{x}_{t,i} \odot \epsilon_\theta(\mathbf{x}_{t,i}, t) \right) 
- \mathbb{E} [{\mathbf{x}_t}] \odot \frac{1}{S} \sum_{i=1}^S \epsilon_\theta(\mathbf{x}_{t,i}, t),
    \label{eq:mc}
\end{equation}
where $\odot$ denotes element-wise multiplication, $S$ is the number of Monte Carlo samples, and $\mathbf{x}_{t,i}$ are samples drawn around the current state  $\mathbf{x}_t$.

The entire uncertainty estimation pipeline operates during inference without any model retraining. The final $\operatorname{Var}(\mathbf{x}_{0})$ provides the pixel-wise uncertainty estimates $\operatorname{Var}(\mbfhat{I})$ for the generated images $\mbfhat{I}$. 
The accuracy of the uncertainty estimates depends on the number of samples and diffusion steps.
In practice, we use $S=20$ samples per step across 50 denoising steps to balance the quality of the uncertainty estimates and computational cost.

\paragraph{Real-world Alignment.}
The DM-generated multi-view images $\mbfhat{I}_{1:k}$ typically lack the correct scale relative to the real-world observations $\mbc{I}$. 
To recover the true scale and align the views to the real-world measurements, we feed the diffusion images to the VGGT network~\cite{wang2025vggt}, which outputs pointcloud $\mbfhat{P}_{1:k}$, associated confidence maps $\mbfhat{C}_{1:k}$, and the relative transformations $\T{C_0}{C_i}$ between input views. 
The predicted confidence maps are modulated by the estimated diffusion uncertainties $\operatorname{Var}(\mbfhat{I})$ through weighted exponential operation $\mbf{C}_{1:k} = \exp \left( \frac{\mbfhat{C}_{1:k}}{\operatorname{Var}(\mbfhat{I}_{1:k})}\right).$

To align the diffusion pointcloud $\mbfhat{P}_{1:k}$ with the real-world pointcloud $\mbf{P}_{0}$ extracted from $\mbc{I}_0$, we first match the eigenvalues of their covariance matrices via Principal Component Analysis (PCA). After recovering the scale $s$, we further refine the rigid transformation between them using Iterative Closest Point (ICP)~\cite{zhang2021fast}. 
The refined diffusion multiview frames together with the first real frame creates an initial pose graph, yielding a metrically consistent 3D reconstruction, aligned with the real-world coordinate frame $\cframe{W}$ while preserving uncertainty estimates.

\subsection{3D Gaussian Splatting Mapping}
\label{sec:gau_map}

With the poses optimized, we proceed to generate a compact 3DGS field~\cite{3dgs} for each object.
Inspired by SplaTAM~\cite{keetha2024splatam}, we model each 3D Gaussian as isotropic, encoding only RGB, 3D position, a scalar radius, and opacity. %
Importantly, we incorporate pixel-wise uncertainty to guide the updates of the 3DGS field when more observations are received.
The overall mapping loss is defined as:
\begin{equation}
    L_t = \sum_{\mbf{p}} \mbf{C}(\mbf{p}) \left( \mbf{S}(\mbf{p}) > 0.99 \right) 
\left( \mathcal{L}_1(D(\mbf{p})) + 0.5 \, \mathcal{L}_1(C(\mbf{p})) \right),
\end{equation}
where $\mbf{C}(\mbf{p})$ denotes the uncertainty at pixel $\textbf{p}$, $\mbf{S}(\mbf{p})$ represents the visibility score, and $D(\mbf{p})$, $C(\mbf{p})$ are the rendered depth and color residual, respectively.
By incorporating certainties from diffusion priors, our mapping is able to always maintain a complete shape while continuously improving he geometric accuracy and photometric consistency when more information are observed.

\begin{figure}[tp]
    \begin{subfigure}[b]{0.48\linewidth}
        \centering
        \includegraphics[width=\linewidth]{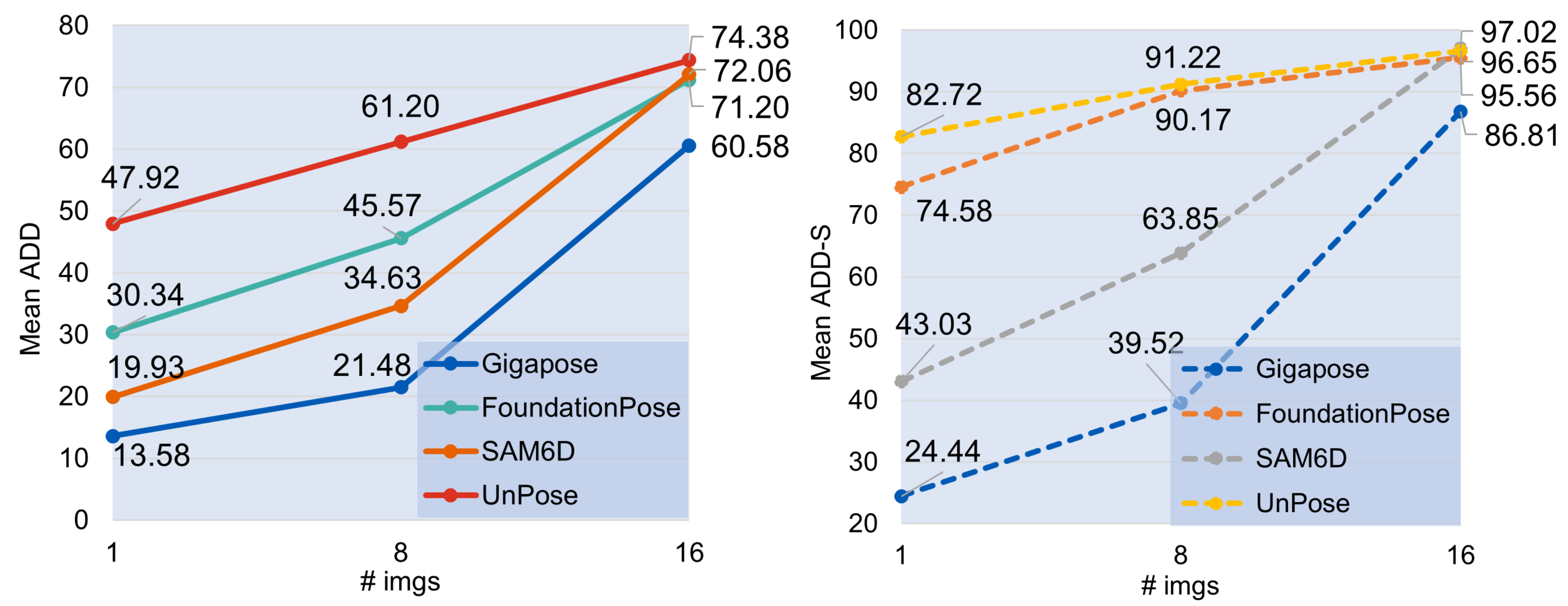}
        \caption{Performance on YCB-Video~\cite{ycbv} dataset.}
        \label{fig:ycbv_pose_trend}
  \end{subfigure}
\hfill
  \begin{subfigure}[b]{0.48\textwidth}
        \centering
        \includegraphics[width=\linewidth]{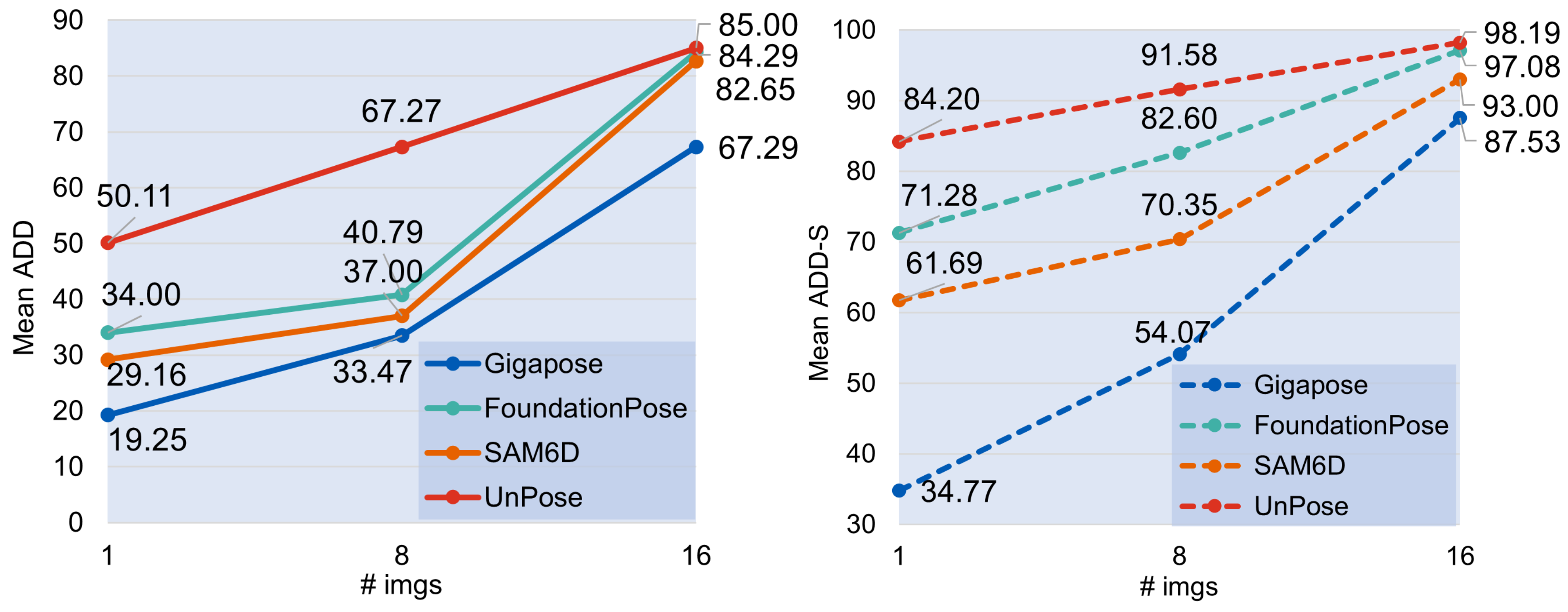}
        \caption{Performance on LM-O~\cite{lmo} dataset.}
        \label{fig:lmo_pose_trend}
  \end{subfigure}
  \caption{We show quantitative comparison on YCB-V~\cite{ycbv} and LM-O~\cite{lmo} using ADD and ADD-S metric \wrt the number of reference-views. \method significantly outperforms the baselines. Note the consistent improvement in performance of our method as the number of reference views increases.}
    \label{fig:pose_results}
\end{figure}

\subsection{6D Pose Estimation}
\label{sec:pose}

Once we have the optimized 3DGS representation of the object, we employ the Pose Refinement network from FoundationPose~\cite{fp} for 6D pose estimation due to its SOTA performance on speed and accuracy.
The network takes two inputs: (1) a rendered RGB-D view  of the object from the current pose estimate (2) a cropped RGB-D observation from the camera. In our implementation, we modify the first input to be rendered directly from the 3DGS field~\cite{3dgs}, rather than a pre-reconstructed 3D model, improving applicability in open-world deployments.
Both inputs are processed through a Siamese encoder with shared weights to extract feature maps, which are concatenated and tokenized into patches with positional embeddings.
The transformer iteratively refines the pose estimates, minimizing the discrepancy between rendered and observed views in feature space.

\subsection{Backend Optimization}
\label{sec:obs_update}
After estimating the relative transformation $\T{O}{C_i}$ for a new frame $\cframe{C_i}$, we evaluate its correspondence to the closest existing keyframe, including both real and virtual diffusion frames, using Mast3R~\cite{mast3r}. 
If the number of matching inliers is below a threshold, the frame is added as a new keyframe $\mathcal{K}_i$ and a bidirectional edge is created between $\mathcal{K}_i$ and its closest keyframe $\mathcal{K}_{i-1}$, updating the pose graph edge set $\mathcal{E}$. 
To close loops for local and global pose graphs, we encode keyframe features using Aggregated Selective Match Kernels (ASMK)~\cite{TAJ13, murai2024_mast3rslam}.

\paragraph{Geometric Pose Graph Optimization.}
Once loop closures are identified, we optimize the geometric pose graph by minimizing a weighted geometric residual over keyframe correspondences:
\begin{equation}
    E_g = \sum_{i,j \in \mathcal{E}} \sum_{m,n \in \mathbf{m}_{i,j}} 
\left\|
\frac{
 \tilde{\mathbf{X}}_{i,m}^i 
- \mathbf{T}_{ij} \tilde{\mathbf{X}}_{j,n}^j
}{
w(\mathbf{q}_{m,n}, \mbf{C}_{m,n}, \sigma_g^2)
}\right\|_{\rho},~
w(\mathbf{q}, \mbf{C}, \sigma^2) = 
\begin{cases}
{\sigma^2/\mathbf{q}} & \text{otherwise} \\ 
\infty & \mathbf{q} < \mathbf{q}_{min} \| \mbf{C} < c_{min}\\
\end{cases}
\end{equation}
where $\mathbf{m}$ denotes set of matches between keyframes $\mathcal{K}_i$ and $\mathcal{K}_j$ predicted by Mast3R~\cite{mast3r}, and $ \tilde{\mathbf{X}}_{i,m}^i$, $\tilde{\mathbf{X}}_{j,n}^j$ are the corresponding 3D points. $\mathbf{q}_{m,n}$ is the match confidence, $\left\|\right\|_{\rho}$ is Huber loss and $w(\mathbf{q}, \mbf{C}, \sigma^2)$ is per-match weighting~\cite{murai2024_mast3rslam} which we extend to include diffusion uncertainties. 
This nonlinear least squares problem is solved efficiently via Gauss-Newton optimization with sparse Cholesky decomposition in CUDA. 
To avoid ambiguity between virtual and real frames, we fix the poses of all diffusion-rendered frames. The refined poses of real keyframes are then sent back to the 3DGS mapping module (see \cref{sec:gau_map}) to ensure a multi-view consistent object representation.

\paragraph{Relocalization.}

If object tracking fails due to occlusion, fast motion or other disruptions, relocalization is triggered. 
Failure is detected through two signals: (a) a sharp drop in frame-to-keyframe correspondence, and (b) a large geometric discrepancy between the observed point cloud and the expected model rendered from the current pose estimate.
Our system benefits from the completeness of the 3D model, allowing even unobserved parts to serve as weak priors.
Upon failure detection, the relocalization module retrieves top-$k$ candidate frames from the retrieval database using ASMK features and ranks them based on Euclidean distance. 
The selected frames are jointly optimized with the current frame to recover its pose, allowing tracking to resume reliably. 

\section{Experiments}
\label{sec:exp}

We evaluate our proposed method,~\method, 
on two benchmarks: YCB-Video~\cite{ycbv} and LM-O~\cite{lmo}.
For pose estimation, we compare against three SOTA baselines: GigaPose~\cite{gigapose} (model-based), SAM6D~\cite{sam6d} (zero-shot reference view-based) and FoundationPose~\cite{fp} (model-free reconstruction-based). As mentioned in~\cref{sec:relatedworks}, these methods require estimation of the complete geometry prior to estimating the 6D pose.
To quantitatively evaluate how the view number of observations impact the pose estimation results, we report the performance given 1, 8, and 16 reference images surrounding the target object, representing pose estimation performance at single view, partial observations, and complete observations cases.
For reconstruction quality, we compare \method against three SOTA baselines: BundleSDF~\cite{bundlesdf} (neural implicit reconstruction), GOM~\cite{gom} (3D diffusion priors-based multi-view optimization), and Wonder3D~\cite{wonder3d} (single-view diffusion-based 3D generation), on several objects from YCB-Video~\cite{ycbv} dataset. 
Similarly, we test the reconstruction performance under 1, 8, 16 reference views.
Pose estimation accuracy is evaluated using 
Area Under the Curve (AUC) of Average
Distance of Model Points (ADD) and its symmetric variant (ADD-S).
Reconstruction quality is measured by Chamfer Distance (CD) to ground truth.
Furthermore, we report the computation time of \method.

\subsection{Results}

\paragraph{Pose Estimation.}

As visible in~\cref{fig:lmo_pose_trend,fig:ycbv_pose_trend}, our method consistently outperforms all baselines in both sparse and dense observation settings, with performance continuously improving as the number of reference views increase.
In the single view case, \method surpasses the model-based GigaPose~\cite{gigapose} by $71.66\%$, and the model-free FoundationPose (also the second-best method)~\cite{fp} by $36.6\%$ on average, benefiting from strong 3D diffusion priors in the initial pose graph.
With additional views, our geometric pose graph optimization enforces multi-view consistency and leverages uncertainty estimates to refine unreliable regions, further enhancing performance.
For the sake of page limit, we show per-object quantitative and qualitative comparisons in the appendix. %

\begin{figure}[htbp]

        \centering
    \includegraphics[width=\linewidth]{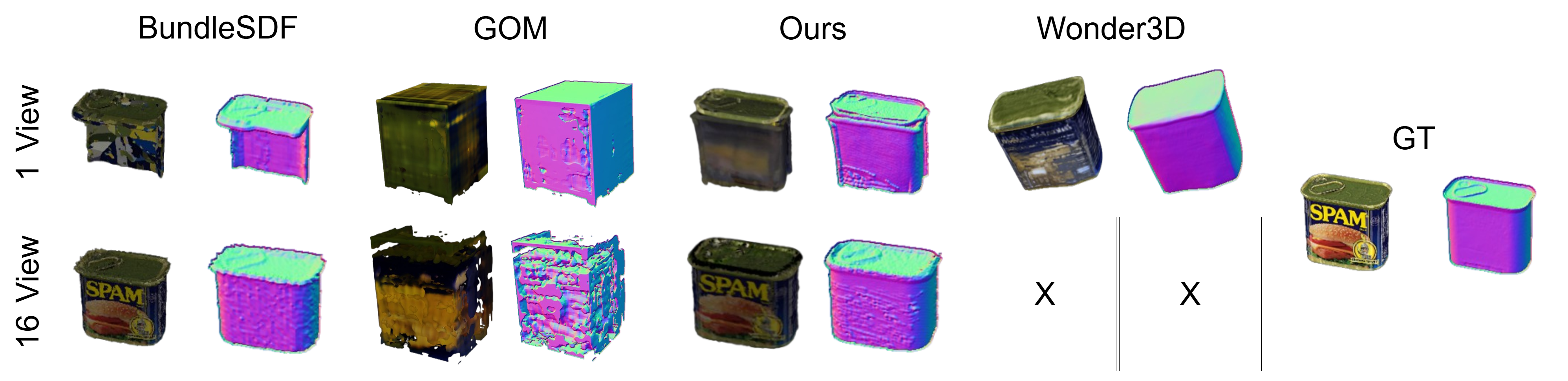}
    \caption{We show qualitative results using our uncertainty-aware 3D reconstruction when 1 and 16 reference views are used for reconstruction. Notice the improvement in geometry and appearance of the object with our method as the number of reference views increases. Since Wonder3D~\cite{wonder3d} does not support multi-view reconstruction, we do not show it. }
    \label{fig:recon}
\end{figure}

\paragraph{Reconstruction.}

\begin{figure}[htbp]
  \centering
  \begin{subfigure}[b]{0.48\linewidth}
    \centering
    \includegraphics[width=0.85\textwidth]{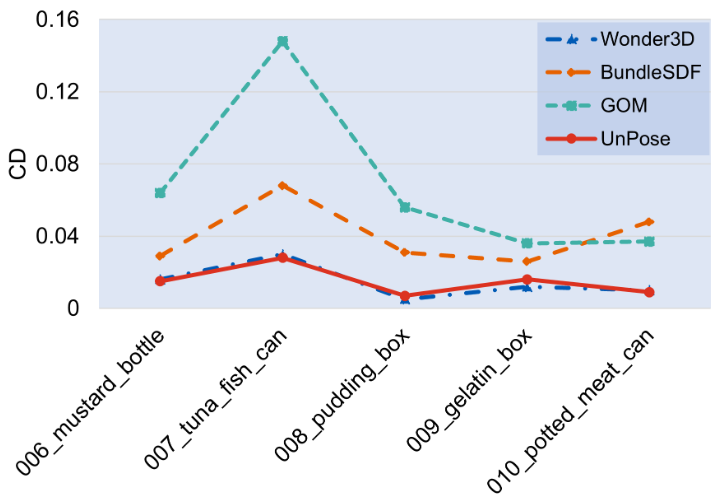}
    \caption{Chamfer Distance}
    \label{fig:cd}
  \end{subfigure}
  \hfill
  \begin{subfigure}[b]{0.48\linewidth}
    \centering
    \includegraphics[width=0.85\textwidth]{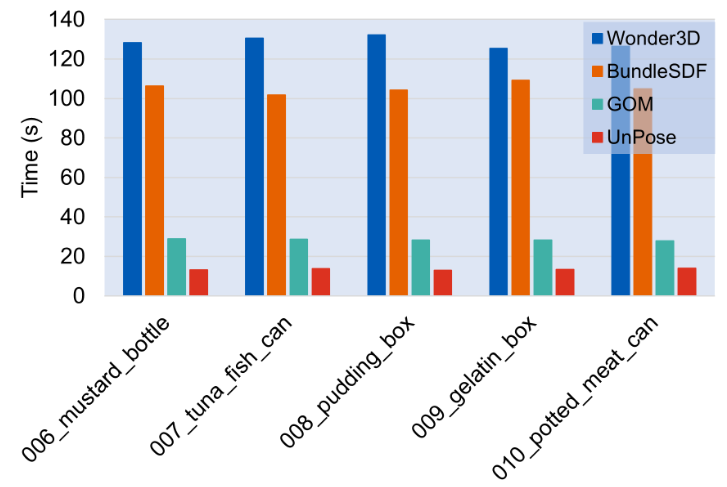}
    \caption{Reconstruction Time (s)}
    \label{fig:time}
  \end{subfigure}
  \caption{Quantitative comparison of object reconstruction for fidelity (a) and efficiency (b) using a single reference view, on objects from YCB-Video~\cite{ycbv} dataset.}
  \label{fig:runtime}
\end{figure}

\cref{fig:runtime} shows the comparison of quantitative results for single-view object reconstruction against baselines~\cite{bundlesdf,gom,wonder3d} on a subset of objects from YCB-Video~\cite{ycbv} dataset.
As visible in~\cref{fig:cd,fig:recon},~\method significantly outperforms the baselines.
Compared to the multiview-to-3D diffusion model~\cite{wonder3d,neus}, \method is significantly faster while yielding comparable reconstruction.
In comparison to other optimization-based methods~\cite{bundlesdf,gom},~\method is $7\times$ and $2\times$ faster while being $2.3\times$ and $3.8\times$ accurate on average, respectively.
These performance gains are owing to the proposed uncertainty-guided reconstruction using 3D diffusion priors and global consistency brought by pose graph optimization.

\paragraph{Runtime Analysis.} 

For a fair comparison, all quantitative comparisons were performed on the same hardware platform.
Further, we report the time for reconstruction compared to prior works in~\cref{fig:time}, showcasing the efficiency and efficacy of our uncertainty-guided diffusion priors.
We also provide a comprehensive analysis of the runtime of each component of our framework in~\cref{tab:runtime}.

\begin{table}[h]
  \centering
  \setlength\tabcolsep{0pt}
  \begin{minipage}{.60\textwidth}
    \large
    \resizebox{\linewidth}{!}{
    \centering
        \begin{threeparttable}
            \begin{tabular}{lcc|cc}
                    \toprule
                    & \multicolumn{2}{c}{16 images}  & \multicolumn{2}{c}{8 images}  \\
                    \cmidrule{2-5}
                      & PSNR $(\uparrow)$ & CD $(\uparrow)$ & PSNR $(\uparrow)$ & CD $(\uparrow)$ \\
                    \midrule
                    Ours & \textbf{35.70} &\textbf{ 0.023} & \textbf{32.90} & \textbf{0.032} \\
                    $\hookrightarrow$ w/o Uncertainty Estimation & 29.80 & 0.027 & 29.3 & 0.034 \\
                    $\hookrightarrow$ w/o Bundle Adjustment & 26.11 & 0.035 & 30.1 & 0.038 \\
                    $\hookrightarrow$ w/o Diffused frame in the pose graph & 35.5 & 0.024 & 28.5 & 0.046 \\
                    \bottomrule
                    \end{tabular}
                \caption{Ablation study demonstrating the impact of uncertainty estimation and incremental refinement on appearance of real views (PSNR) and reconstructed geometry (CD). $\uparrow$ denotes higher is better.}
                \label{tab:ablation}
        \end{threeparttable}
        }
  \end{minipage}
  \hspace{1em}
  \begin{minipage}{.32\textwidth}
    \centering
    \large
    \resizebox{\linewidth}{!}{
    \begin{threeparttable}
        \begin{tabular}{lc}
            \toprule
            Process & Time (s) \\
            \midrule
            Uncertainty Estimation & 12.08 \\
            Remaining Initialization (\cref{sec:init}) & 1.07 \\
            3DGS Mapping (\cref{sec:gau_map}) & 0.22 \\
            Pose Estimation (\cref{sec:pose}) & 0.12 \\
            \cmidrule{1-2}
            Backend Optimization (\cref{sec:obs_update}) & 1.70 \\
            \midrule
            Total & 13.49  \\
            \bottomrule
        \end{tabular}
        \caption{Runtime of each component of our pipeline.}
        \label{tab:runtime}
        \end{threeparttable}
    }
  \end{minipage}
\end{table}

\subsection{Ablation Study}

We tested our method performance without  uncertainty information, without bundle adjustment, and without including diffused frames in pose graph optimization in two scenarios: 16 images and 8 images, using PSNR and CD metrics. With 16 images, uncertainty guidance and bundle adjustment improved the performance by 20\%, while incorporating diffused frames had minimum effect when the density of observations is sufficient. In contrast, with only 8 images, incorporating diffused frames into the pose graph yielded a 15\% performance gain, highlighting their importance in sparser observation scenarios.

\subsection{Robotic Application}
We further demonstrate the effectiveness of~\method in robot arm manipulation tasks, where it is used to estimate an object's 6D pose and completed geometry,
followed by grasp pose estimation using AnyGrasp~\cite{fang2023anygrasp}.
An example can be seen in~\cref{fig:robot}, the in-hand camera initially captures the mug with its handle occluded, leading to a grasp near the cup's wall. This is not ideal for tasks such as water pouring. 
As new views reveal the handle, \method incrementally updates the geometry, resulting in 
a new handle-aligned grasp pose, thereby allowing the robot arm to successfully grasp the mug by its handle. More details are provided in \cref{app:real_world_robot}. 

\section{Conclusion}
\label{sec:conclusion}

We presented~\method, a novel framework for model-free 6D pose estimation that exploits 3D priors and uncertainty estimates from a diffusion model.
Our method incrementally refines object geometry by fusing diffused views with their uncertainty estimates and incorporating them into a factor graph.
We demonstrated qualitatively and quantitatively the advantages of our proposed method for pose estimation and reconstruction tasks.%

\clearpage
\section{Limitations}
Our current backend optimization relies on correspondence computed from Mast3R\cite{mast3r}, which limits performance on textureless objects. A promising direction is to integrate direct matching losses from the direct SLAM literature~\cite{whelan2016elasticfusion}. Additionally, despite the significant performance gains, real-time deployment remains a challenge due to the computational overhead of Monte Carlo sampling for diffusion-based uncertainty estimation. This could be addressed by training a diffusion model that jointly predicts both mean and uncertainty. In addition, our current focus is on individual objects, incorporating scene-level priors and cross-object relationships may further improve global consistency. We believe these limitation analysis and future directions offer valuable opportunities for the community.

\acknowledgments{The authors would like to thanks the reviewers for their constructive feedback on improving this work. We would also like to the thank the members of Huawei Noah's Ark Lab for insightful discussions during the development of this project.
}

\small
\bibliography{example}  %
\newpage
\appendix
\section{Additional Qualitative Results}

\begin{figure}[h]
    \centering
    \includegraphics[width=\linewidth]{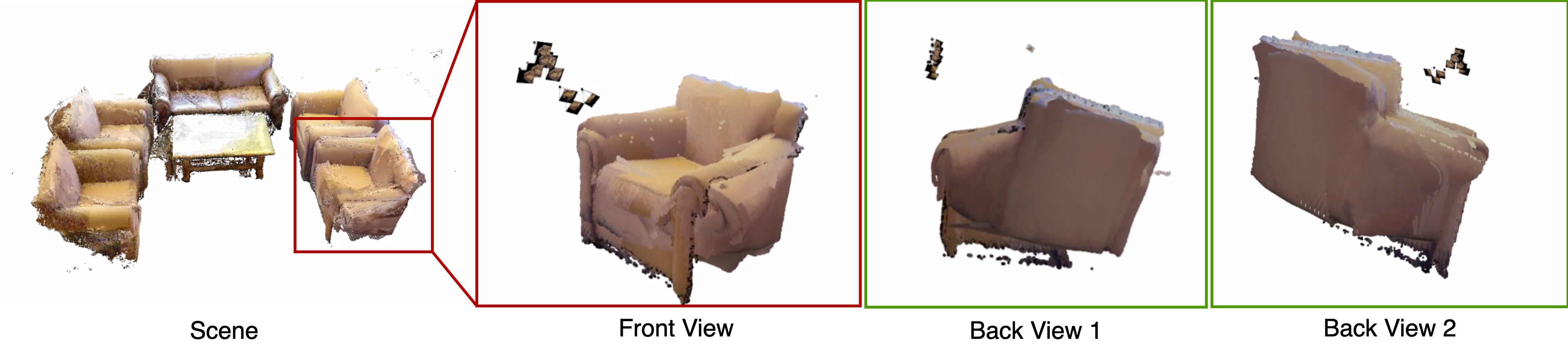}
    \caption{Visualization of a scene-level reconstruction on a ScanNet scene~\cite{scannet} by \method. We show the reconstruction of one of the chairs from one input camera viewpoint (Front View) as well as diffused novel views (Back View 1 and 2). Real and diffusion images of the scene used in backend optimization are shown as floaters.}
    \label{fig:scannet}
\end{figure}

\paragraph{Scene-level Reconstructions.}
To demonstrate the generalization capablities of \method, we test its performance beyond tabletop objects (YCB-V, LM-O) on large-scale furniture objects from the ScanNet~\cite{scannet}.
As shown in~\cref{fig:scannet}, \method reconstruct multiple diverse objects in a scene, including four chairs, one sofa, and one table, all from the same pipeline.
We showcase the reconstruction of one object (chair) in the scene, and show it from an input camera viewpoint (Front View) and two novel views (Back View 1 and 2). The smaller floater images are real observations and diffusion frames provided to backend optimization, offering context on the input data. 

\begin{figure}[h]
    \centering
    \includegraphics[width=\textwidth]{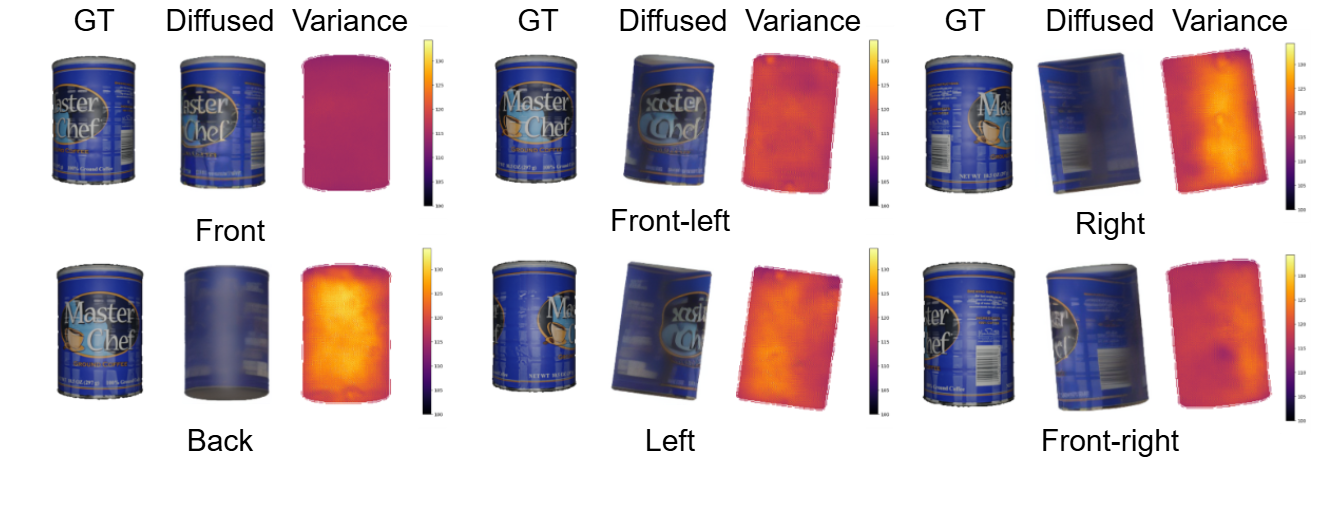}
    \caption{Visualization of diffused views, the associated pixel-level uncertainty estimates, and the corresponding ground-truth rendering perspectives. The diffused images show larger variance at unseen angles.}
    \label{fig:uncertainity_all}
\end{figure}

\cref{fig:uncertainity_all} visualizes the pixel-level uncertainty estimations associated with the novel views generated by the diffusion model~\cite{wonder3d} used in our pipeline. 
For a better understanding, we also show the ground truth renderings of the object alongside each diffused view.
It can be seen that the diffusion model exhibits higher variance (greater uncertainties) for novel views that significantly deviates from the input. This implies the model is less confident when synthesizing unseen regions of the object.

\begin{figure}[h]
    \centering
    \includegraphics[width=\linewidth]{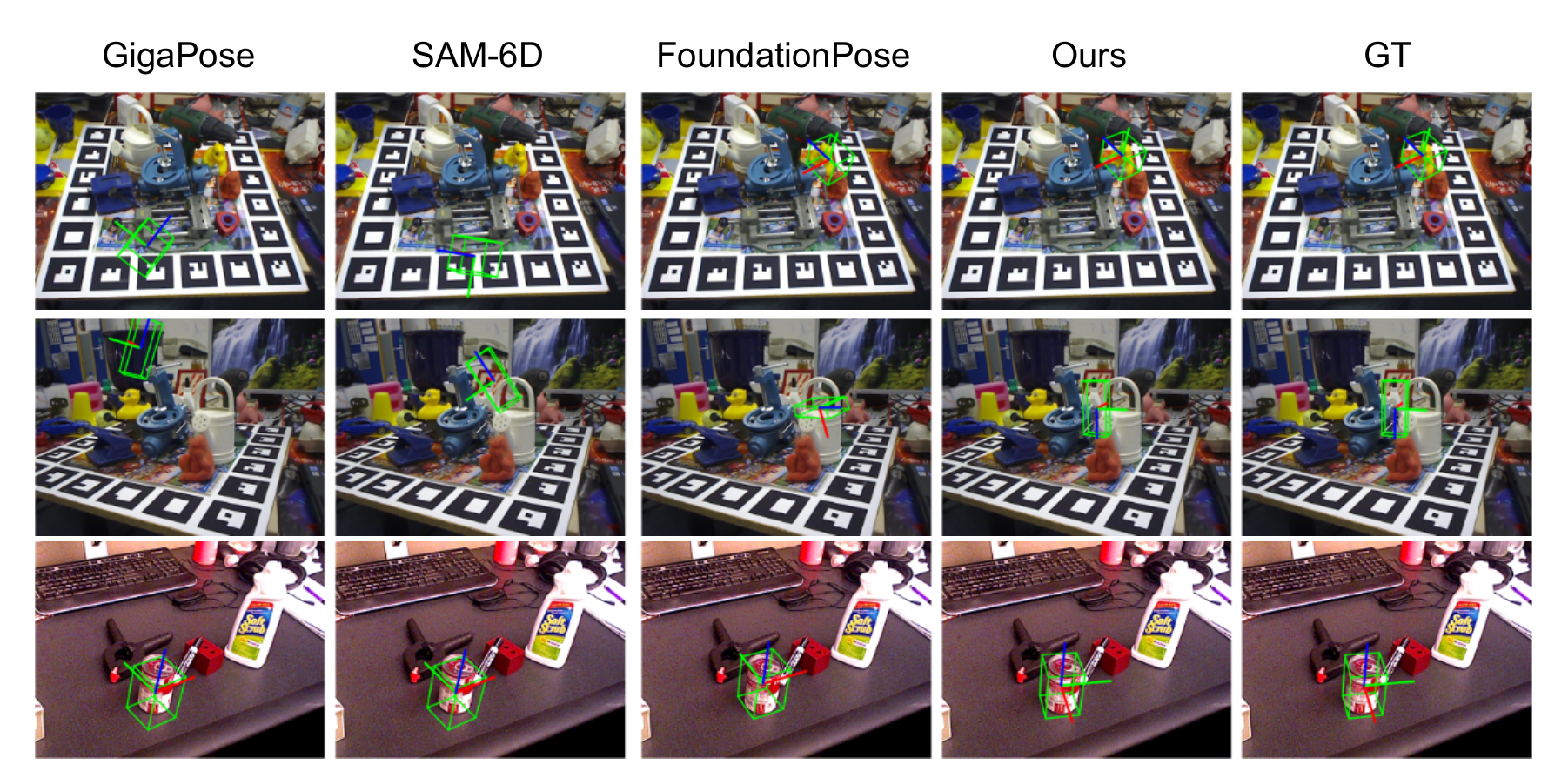}
    \caption{Qualitative results of~\method compared to prior pose estimation methods on YCB-Video~\cite{ycbv} and LM-O~\cite{lmo} datasets when 8 reference views are used.
    }
    \label{fig:pose_ycbv}
\end{figure}

We further provide qualitative comparisons with SOTA 6D pose estimation methods~\cite{gigapose,sam6d,fp}, including GigaPose~\cite{gigapose}, SAM-6D~\cite{sam6d}, and FoundationPose~\cite{fp} on standard 6D pose estimation benchmark datasets: YCB-Video~\cite{ycbv} and LM-O~\cite{lmo}.
 \cref{fig:pose_ycbv} clearly shows that compared to prior methods where the estimated translation and rotation show high error, our proposed method,~\method, accurately estimates the 6DOF of the object.

\section{Additional Quantitative Results}

In this section, we present additional quantitative comparison details on YCB-Video~\cite{ycbv} and LM-O~\cite{lmo} datasets, and further experiments on T-Less~\cite{hodan2017t} and TYO-L~\cite{hodan2018bop}.
We report object reconstruction results on YCB-V~\cite{ycbv} dataset in~\cref{tab:recon_all}, evaluating performance with 1, 8 and 16 images.
Our proposed method consistently achieves lower reconstruction error (measured as Chamfer Distance) and is significantly faster than prior methods~\cite{gom,bundlesdf,wonder3d}.
We tabulate the 6D pose estimation results on YCB-Video~\cite{ycbv}, LM-O~\cite{lmo}, T-Less~\cite{hodan2017t}, and TYO-L~\cite{hodan2018bop} in \cref{tab:ycb_pose,tab:lmo_pose,tab:tless_tyol_split} respectively.
Following prior works~\cite{fp,liu2025hippoharnessingimageto3dpriors,lee2025any6d}, we evaluate the performance on pose estimation using ADD and ADD-S.
Furthermore, for each method we report the results when 1, 8 and 16 images are used for reconstruction~\cite{gigapose,fp} or as reference views~\cite{sam6d}.
It can be seen in~\cref{tab:ycb_pose,tab:lmo_pose,tab:tless_tyol_split} that our method consistently outperforms prior SOTA methods (visible as gradual increase in the brightness of the gradient) in all settings~\ie, both metric and the number of images.

\begin{table}[h]
\centering
\resizebox{\textwidth}{!}{
\Large
\begin{tabular}{c|c|ccc|ccc|ccc|ccc|ccc|ccc}
\toprule
\multirow{2}{*}{Method} & \multirow{2}{*}{Metric}  & \multicolumn{3}{c}{$ycbv_5$} & \multicolumn{3}{c}{$ycbv_6$}   & \multicolumn{3}{c}{$ycbv_7$} &  \multicolumn{3}{c}{$ycbv_8$} &   \multicolumn{3}{c}{$ycbv_9$} & \multicolumn{3}{c}{$ycbv_{10}$}  \\
\cmidrule{3-20}
 &  & 1 & 8 & 16 & 1 & 8 & 16 & 1 & 8 & 16 & 1 & 8 & 16 & 1 & 8 & 16 & 1 & 8 & 16 \\
\midrule
\multirow[c]{2}{*}{Wonder3d~\cite{wonder3d}} & CD & {\cellcolor[HTML]{F7FCF0}} \color[HTML]{000000} 0.016 & {\cellcolor[HTML]{F7FCF0}} \color[HTML]{000000} 0.00 & {\cellcolor[HTML]{F7FCF0}} \color[HTML]{000000} 0.00 & {\cellcolor[HTML]{F7FCF0}} \color[HTML]{000000} 0.030 & {\cellcolor[HTML]{F7FCF0}} \color[HTML]{000000} 0.00 & {\cellcolor[HTML]{F7FCF0}} \color[HTML]{000000} 0.00 & {\cellcolor[HTML]{F7FCF0}} \color[HTML]{000000} 0.005 & {\cellcolor[HTML]{F7FCF0}} \color[HTML]{000000} 0.00 & {\cellcolor[HTML]{F7FCF0}} \color[HTML]{000000} 0.00 & {\cellcolor[HTML]{F7FCF0}} \color[HTML]{000000} 0.012 & {\cellcolor[HTML]{F7FCF0}} \color[HTML]{000000} 0.00 & {\cellcolor[HTML]{F7FCF0}} \color[HTML]{000000} 0.00 & {\cellcolor[HTML]{F7FCF0}} \color[HTML]{000000} 0.010 & {\cellcolor[HTML]{F7FCF0}} \color[HTML]{000000} 0.00 & {\cellcolor[HTML]{F7FCF0}} \color[HTML]{000000} 0.00 & {\cellcolor[HTML]{F7FCF0}} \color[HTML]{000000} 0.021 & {\cellcolor[HTML]{F7FCF0}} \color[HTML]{000000} 0.00 & {\cellcolor[HTML]{F7FCF0}} \color[HTML]{000000} 0.00 \\
 & Time (s) & {\cellcolor[HTML]{084081}} \color[HTML]{F1F1F1} 128.20 & {\cellcolor[HTML]{F7FCF0}} \color[HTML]{000000} 0.00 & {\cellcolor[HTML]{F7FCF0}} \color[HTML]{000000} 0.00 & {\cellcolor[HTML]{084081}} \color[HTML]{F1F1F1} 130.63 & {\cellcolor[HTML]{F7FCF0}} \color[HTML]{000000} 0.00 & {\cellcolor[HTML]{F7FCF0}} \color[HTML]{000000} 0.00 & {\cellcolor[HTML]{084081}} \color[HTML]{F1F1F1} 132.22 & {\cellcolor[HTML]{F7FCF0}} \color[HTML]{000000} 0.00 & {\cellcolor[HTML]{F7FCF0}} \color[HTML]{000000} 0.00 & {\cellcolor[HTML]{084081}} \color[HTML]{F1F1F1} 125.30 & {\cellcolor[HTML]{F7FCF0}} \color[HTML]{000000} 0.00 & {\cellcolor[HTML]{F7FCF0}} \color[HTML]{000000} 0.00 & {\cellcolor[HTML]{084081}} \color[HTML]{F1F1F1} 126.40 & {\cellcolor[HTML]{F7FCF0}} \color[HTML]{000000} 0.00 & {\cellcolor[HTML]{F7FCF0}} \color[HTML]{000000} 0.00 & {\cellcolor[HTML]{084081}} \color[HTML]{F1F1F1} 130.20 & {\cellcolor[HTML]{F7FCF0}} \color[HTML]{000000} 0.00 & {\cellcolor[HTML]{F7FCF0}} \color[HTML]{000000} 0.00 \\
\cmidrule{1-20}
\multirow[c]{2}{*}{BundleSDF~\cite{bundlesdf}} & CD & {\cellcolor[HTML]{F7FCF0}} \color[HTML]{000000} 0.029 & {\cellcolor[HTML]{F7FCF0}} \color[HTML]{000000} 0.020 & {\cellcolor[HTML]{F7FCF0}} \color[HTML]{000000} 0.016 & {\cellcolor[HTML]{F7FCF0}} \color[HTML]{000000} 0.068 & {\cellcolor[HTML]{F7FCF0}} \color[HTML]{000000} 0.044 & {\cellcolor[HTML]{F7FCF0}} \color[HTML]{000000} 0.015 & {\cellcolor[HTML]{F7FCF0}} \color[HTML]{000000} 0.031 & {\cellcolor[HTML]{F7FCF0}} \color[HTML]{000000} 0.017 & {\cellcolor[HTML]{F7FCF0}} \color[HTML]{000000} 0.011 & {\cellcolor[HTML]{F7FCF0}} \color[HTML]{000000} 0.026 & {\cellcolor[HTML]{F7FCF0}} \color[HTML]{000000} 0.018 & {\cellcolor[HTML]{F7FCF0}} \color[HTML]{000000} 0.011 & {\cellcolor[HTML]{F7FCF0}} \color[HTML]{000000} 0.048 & {\cellcolor[HTML]{F7FCF0}} \color[HTML]{000000} 0.024 & {\cellcolor[HTML]{F7FCF0}} \color[HTML]{000000} 0.014 & {\cellcolor[HTML]{F7FCF0}} \color[HTML]{000000} 0.005 & {\cellcolor[HTML]{F7FCF0}} \color[HTML]{000000} 0.004 & {\cellcolor[HTML]{F7FCF0}} \color[HTML]{000000} 0.001 \\
 & Time (s) & {\cellcolor[HTML]{1475B2}} \color[HTML]{F1F1F1} 106.35 & {\cellcolor[HTML]{084081}} \color[HTML]{F1F1F1} 135.27 & {\cellcolor[HTML]{084081}} \color[HTML]{F1F1F1} 172.96 & {\cellcolor[HTML]{2283BA}} \color[HTML]{F1F1F1} 101.74 & {\cellcolor[HTML]{084081}} \color[HTML]{F1F1F1} 112.90 & {\cellcolor[HTML]{084586}} \color[HTML]{F1F1F1} 126.31 & {\cellcolor[HTML]{2081B8}} \color[HTML]{F1F1F1} 104.23 & {\cellcolor[HTML]{084081}} \color[HTML]{F1F1F1} 118.74 & {\cellcolor[HTML]{084081}} \color[HTML]{F1F1F1} 131.49 & {\cellcolor[HTML]{0868AC}} \color[HTML]{F1F1F1} 109.20 & {\cellcolor[HTML]{084081}} \color[HTML]{F1F1F1} 114.46 & {\cellcolor[HTML]{084081}} \color[HTML]{F1F1F1} 130.76 & {\cellcolor[HTML]{1475B2}} \color[HTML]{F1F1F1} 104.97 & {\cellcolor[HTML]{084081}} \color[HTML]{F1F1F1} 129.13 & {\cellcolor[HTML]{084081}} \color[HTML]{F1F1F1} 147.39 & {\cellcolor[HTML]{2182B9}} \color[HTML]{F1F1F1} 102.11 & {\cellcolor[HTML]{084081}} \color[HTML]{F1F1F1} 115.84 & {\cellcolor[HTML]{084081}} \color[HTML]{F1F1F1} 132.76 \\
\cmidrule{1-20}
\multirow[c]{2}{*}{GOM~\cite{gom}} & CD & {\cellcolor[HTML]{F7FCF0}} \color[HTML]{000000} 0.064 & {\cellcolor[HTML]{F7FCF0}} \color[HTML]{000000} 0.117 & {\cellcolor[HTML]{F7FCF0}} \color[HTML]{000000} 0.080 & {\cellcolor[HTML]{F7FCF0}} \color[HTML]{000000} 0.148 & {\cellcolor[HTML]{F7FCF0}} \color[HTML]{000000} 0.121 & {\cellcolor[HTML]{F7FCF0}} \color[HTML]{000000} 0.100 & {\cellcolor[HTML]{F7FCF0}} \color[HTML]{000000} 0.056 & {\cellcolor[HTML]{F7FCF0}} \color[HTML]{000000} 0.066 & {\cellcolor[HTML]{F7FCF0}} \color[HTML]{000000} 0.079 & {\cellcolor[HTML]{F7FCF0}} \color[HTML]{000000} 0.036 & {\cellcolor[HTML]{F7FCF0}} \color[HTML]{000000} 0.169 & {\cellcolor[HTML]{F7FCF0}} \color[HTML]{000000} 0.092 & {\cellcolor[HTML]{F7FCF0}} \color[HTML]{000000} 0.037 & {\cellcolor[HTML]{F7FCF0}} \color[HTML]{000000} 0.039 & {\cellcolor[HTML]{F7FCF0}} \color[HTML]{000000} 0.046 & {\cellcolor[HTML]{F7FCF0}} \color[HTML]{000000} 0.153 & {\cellcolor[HTML]{F7FCF0}} \color[HTML]{000000} 0.268 & {\cellcolor[HTML]{F7FCF0}} \color[HTML]{000000} 0.170 \\
 & Time (s) & {\cellcolor[HTML]{D0EDCA}} \color[HTML]{000000} 28.88 & {\cellcolor[HTML]{A0DAB8}} \color[HTML]{000000} 53.64 & {\cellcolor[HTML]{389BC6}} \color[HTML]{F1F1F1} 121.19 & {\cellcolor[HTML]{D1EDCA}} \color[HTML]{000000} 28.63 & {\cellcolor[HTML]{80CEC2}} \color[HTML]{000000} 54.92 & {\cellcolor[HTML]{084081}} \color[HTML]{F1F1F1} 128.65 & {\cellcolor[HTML]{D2EDCC}} \color[HTML]{000000} 28.38 & {\cellcolor[HTML]{8DD3BE}} \color[HTML]{000000} 53.53 & {\cellcolor[HTML]{085A9D}} \color[HTML]{F1F1F1} 120.24 & {\cellcolor[HTML]{D0EDCA}} \color[HTML]{000000} 28.39 & {\cellcolor[HTML]{8AD2BF}} \color[HTML]{000000} 52.47 & {\cellcolor[HTML]{0861A4}} \color[HTML]{F1F1F1} 116.98 & {\cellcolor[HTML]{D1EDCA}} \color[HTML]{000000} 27.97 & {\cellcolor[HTML]{9BD8B9}} \color[HTML]{000000} 53.11 & {\cellcolor[HTML]{2182B9}} \color[HTML]{F1F1F1} 115.53 & {\cellcolor[HTML]{D1EDCB}} \color[HTML]{000000} 28.27 & {\cellcolor[HTML]{8BD2BF}} \color[HTML]{000000} 52.58 & {\cellcolor[HTML]{0866A9}} \color[HTML]{F1F1F1} 117.18 \\
\cmidrule{1-20}
\multirow[c]{2}{*}{Ours} & CD & {\cellcolor[HTML]{F7FCF0}} \color[HTML]{000000} 0.015 & {\cellcolor[HTML]{F7FCF0}} \color[HTML]{000000} 0.005 & {\cellcolor[HTML]{F7FCF0}} \color[HTML]{000000} 0.004 & {\cellcolor[HTML]{F7FCF0}} \color[HTML]{000000} 0.028 & {\cellcolor[HTML]{F7FCF0}} \color[HTML]{000000} 0.023 & {\cellcolor[HTML]{F7FCF0}} \color[HTML]{000000} 0.020 & {\cellcolor[HTML]{F7FCF0}} \color[HTML]{000000} 0.007 & {\cellcolor[HTML]{F7FCF0}} \color[HTML]{000000} 0.006 & {\cellcolor[HTML]{F7FCF0}} \color[HTML]{000000} 0.005 & {\cellcolor[HTML]{F7FCF0}} \color[HTML]{000000} 0.016 & {\cellcolor[HTML]{F7FCF0}} \color[HTML]{000000} 0.008 & {\cellcolor[HTML]{F7FCF0}} \color[HTML]{000000} 0.007 & {\cellcolor[HTML]{F7FCF0}} \color[HTML]{000000} 0.009 & {\cellcolor[HTML]{F7FCF0}} \color[HTML]{000000} 0.009 & {\cellcolor[HTML]{F7FCF0}} \color[HTML]{000000} 0.004 & {\cellcolor[HTML]{F7FCF0}} \color[HTML]{000000} 0.021 & {\cellcolor[HTML]{F7FCF0}} \color[HTML]{000000} 0.005 & {\cellcolor[HTML]{F7FCF0}} \color[HTML]{000000} 0.003 \\
 & Time (s) & {\cellcolor[HTML]{E4F5DF}} \color[HTML]{000000} 13.20 & {\cellcolor[HTML]{D5EFCF}} \color[HTML]{000000} 26.20 & {\cellcolor[HTML]{CEECC8}} \color[HTML]{000000} 41.20 & {\cellcolor[HTML]{E4F5DF}} \color[HTML]{000000} 13.80 & {\cellcolor[HTML]{CEECC7}} \color[HTML]{000000} 27.20 & {\cellcolor[HTML]{B7E3BC}} \color[HTML]{000000} 41.70 & {\cellcolor[HTML]{E6F5E0}} \color[HTML]{000000} 12.90 & {\cellcolor[HTML]{D1EDCA}} \color[HTML]{000000} 26.30 & {\cellcolor[HTML]{B5E2BB}} \color[HTML]{000000} 43.22 & {\cellcolor[HTML]{E4F4DE}} \color[HTML]{000000} 13.50 & {\cellcolor[HTML]{D1EDCA}} \color[HTML]{000000} 25.30 & {\cellcolor[HTML]{BCE5BE}} \color[HTML]{000000} 40.25 & {\cellcolor[HTML]{E3F4DE}} \color[HTML]{000000} 14.10 & {\cellcolor[HTML]{D8F0D3}} \color[HTML]{000000} 22.21 & {\cellcolor[HTML]{C4E8C1}} \color[HTML]{000000} 41.30 & {\cellcolor[HTML]{E7F6E2}} \color[HTML]{000000} 11.30 & {\cellcolor[HTML]{D4EECE}} \color[HTML]{000000} 23.20 & {\cellcolor[HTML]{B9E3BC}} \color[HTML]{000000} 42.10 \\
\bottomrule
\end{tabular}
}
\caption{Quantitative comparison for reconstruction with $1, 8$ and $16$ images using~\method in terms of accuracy (reported using chamfer distance (CD) as $10^{-3}$) and time (sec) on objects from YCB-V~\cite{ycbv}. For both CD and time, brighter gradients denotes better performance.}
\label{tab:recon_all}
\end{table}

\begin{table}[htbp]
  \centering
  \setlength{\tabcolsep}{4pt} %
  \Large

  \begin{minipage}{0.49\textwidth}
    \centering
    \resizebox{\linewidth}{!}{%
      \begin{tabular}{c c cccc}
        \multicolumn{6}{c}{\textbf{(a) T-LESS~\cite{hodan2017t}}} \\
        \toprule
        \# imgs & Metric & GigaPose~\cite{gigapose} & SAM6D~\cite{sam6d} & FoundationPose~\cite{fp} & Ours \\
        \midrule
        \multirow{2}{*}{1}
          & ADD   & 10.0 & 14.9 & \underline{16.6} & \textbf{32.5} \\
          & ADD-S & 71.3 & 78.5 & \underline{82.2} & \textbf{89.2} \\
        \midrule
        \multirow{2}{*}{8}
          & ADD   & 17.2 & 28.9 & \underline{33.2} & \textbf{40.1} \\
          & ADD-S & 73.0 & 81.8 & \underline{85.4} & \textbf{93.6} \\
        \midrule
        \multirow{2}{*}{16}
          & ADD   & 44.0 & 50.1 & \underline{53.5} & \textbf{54.2} \\
          & ADD-S & 76.7 & 89.5 & \underline{90.7} & \textbf{94.7} \\
        \bottomrule
      \end{tabular}
    }
  \end{minipage}%
  \hfill
  \begin{minipage}{0.49\textwidth}
    \centering
    \resizebox{\linewidth}{!}{%
      \begin{tabular}{c c cccc}
        \multicolumn{6}{c}{\textbf{(b) TYO-L~\cite{hodan2018bop}}} \\
        \toprule
        \# imgs & Metric & GigaPose~\cite{gigapose} & SAM6D~\cite{sam6d} & FoundationPose~\cite{fp} & Ours \\
        \midrule
        \multirow{2}{*}{1}
          & ADD   & 14.2 & 37.5 & \underline{42.2} & \textbf{55.3} \\
          & ADD-S & 62.2 & 75.8 & \underline{82.7} & \textbf{88.2} \\
        \midrule
        \multirow{2}{*}{8}
          & ADD   & 23.3 & 42.7 & \underline{45.5} & \textbf{67.2} \\
          & ADD-S & 72.2 & 82.6 & \underline{86.2} & \textbf{93.6} \\
        \midrule
        \multirow{2}{*}{16}
          & ADD   & 59.2 & 68.3 & \underline{72.2} & \textbf{75.3} \\
          & ADD-S & 87.3 & 93.3 & \underline{97.3} & \textbf{98.1} \\
        \bottomrule
      \end{tabular}
    }
  \end{minipage}

  \caption{Quantitative comparison of pose estimation on T-LESS~\cite{hodan2017t} and TYO-L~\cite{hodan2018bop}. Best in \textbf{bold}; second best is \underline{underlined}.}
  \label{tab:tless_tyol_split}
\end{table}

\begin{table}[h]
    \centering
    \resizebox{\textwidth}{!}{%
    \Large
    \begin{tabular}{c|c|l|*{5}{c|}c}
        \toprule
        Method & \# imgs & Metric & Mean & bath duck & cat toy & hole puncher & power drill & water can \\
        \midrule
        \multirow[c]{6}{*}{Gigapose~\cite{gigapose}} & \multirow[c]{2}{*}{1} & ADD & 19.25 & {\cellcolor[HTML]{084081}} \color[HTML]{F1F1F1} 1.07 & {\cellcolor[HTML]{084081}} \color[HTML]{F1F1F1} 1.15 & {\cellcolor[HTML]{084081}} \color[HTML]{F1F1F1} 2.53 & {\cellcolor[HTML]{084081}} \color[HTML]{F1F1F1} 48.65 & {\cellcolor[HTML]{1576B3}} \color[HTML]{F1F1F1} 42.86 \\
         &  & ADD-S & 34.77 & {\cellcolor[HTML]{085395}} \color[HTML]{F1F1F1} 6.82 & {\cellcolor[HTML]{085395}} \color[HTML]{F1F1F1} 6.90 & {\cellcolor[HTML]{085497}} \color[HTML]{F1F1F1} 8.63 & {\cellcolor[HTML]{BAE4BD}} \color[HTML]{000000} 83.78 & {\cellcolor[HTML]{87D1C0}} \color[HTML]{000000} 67.72 \\
         & \multirow[c]{2}{*}{8} & ADD & 33.46 & {\cellcolor[HTML]{0F6FAF}} \color[HTML]{F1F1F1} 15.56 & {\cellcolor[HTML]{084B8D}} \color[HTML]{F1F1F1} 4.87 & {\cellcolor[HTML]{085A9D}} \color[HTML]{F1F1F1} 10.84 & {\cellcolor[HTML]{47ABCF}} \color[HTML]{F1F1F1} 66.67 & {\cellcolor[HTML]{8FD4BD}} \color[HTML]{000000} 69.39 \\
         &  & ADD-S & 54.07 & {\cellcolor[HTML]{42A6CC}} \color[HTML]{F1F1F1} 33.37 & {\cellcolor[HTML]{0864A8}} \color[HTML]{F1F1F1} 12.43 & {\cellcolor[HTML]{77CAC5}} \color[HTML]{000000} 50.21 & {\cellcolor[HTML]{D8F0D2}} \color[HTML]{000000} 90.90 & {\cellcolor[HTML]{CEECC7}} \color[HTML]{000000} 83.45 \\
         & \multirow[c]{2}{*}{16} & ADD & 67.29 & {\cellcolor[HTML]{AFE0B8}} \color[HTML]{000000} 63.82 & {\cellcolor[HTML]{99D7BA}} \color[HTML]{000000} 57.47 & {\cellcolor[HTML]{3294C2}} \color[HTML]{F1F1F1} 29.36 & {\cellcolor[HTML]{D3EECD}} \color[HTML]{000000} 89.47 & {\cellcolor[HTML]{EEF8E7}} \color[HTML]{000000} 96.36 \\
         &  & ADD-S & 87.53 & {\cellcolor[HTML]{F7FCF0}} \color[HTML]{000000} 97.28 & {\cellcolor[HTML]{F7FCF0}} \color[HTML]{000000} 97.70 & {\cellcolor[HTML]{6CC4C9}} \color[HTML]{000000} 47.37 & {\cellcolor[HTML]{EEF8E7}} \color[HTML]{000000} 97.37 & {\cellcolor[HTML]{F2FAEB}} \color[HTML]{000000} 97.96 \\
        \midrule
        \multirow[c]{6}{*}{FoundationPose~\cite{fp}} & \multirow[c]{2}{*}{1} & ADD & 34.00 & {\cellcolor[HTML]{08599C}} \color[HTML]{F1F1F1} 8.70 & {\cellcolor[HTML]{2D8FBF}} \color[HTML]{F1F1F1} 26.12 & {\cellcolor[HTML]{1D7EB7}} \color[HTML]{F1F1F1} 22.00 & {\cellcolor[HTML]{B1E0B9}} \color[HTML]{000000} 82.23 & {\cellcolor[HTML]{084081}} \color[HTML]{F1F1F1} 30.95 \\
         &  & ADD-S & 71.28 & {\cellcolor[HTML]{47ABCF}} \color[HTML]{F1F1F1} 34.78 & {\cellcolor[HTML]{D4EECE}} \color[HTML]{000000} 78.10 & {\cellcolor[HTML]{BCE5BE}} \color[HTML]{000000} 70.10 & {\cellcolor[HTML]{D5EFCF}} \color[HTML]{000000} 90.13 & {\cellcolor[HTML]{CEECC7}} \color[HTML]{000000} 83.33 \\
         & \multirow[c]{2}{*}{8} & ADD & 40.79 & {\cellcolor[HTML]{1B7BB6}} \color[HTML]{F1F1F1} 19.57 & {\cellcolor[HTML]{2484BA}} \color[HTML]{F1F1F1} 22.78 & {\cellcolor[HTML]{1272B1}} \color[HTML]{F1F1F1} 18.26 & {\cellcolor[HTML]{D0ECC9}} \color[HTML]{000000} 88.23 & {\cellcolor[HTML]{47ABCF}} \color[HTML]{F1F1F1} 55.14 \\
         &  & ADD-S & 82.60 & {\cellcolor[HTML]{D5EECE}} \color[HTML]{000000} 78.27 & {\cellcolor[HTML]{E3F4DE}} \color[HTML]{000000} 86.83 & {\cellcolor[HTML]{B1E0B9}} \color[HTML]{000000} 66.26 & {\cellcolor[HTML]{EAF7E4}} \color[HTML]{000000} 96.33 & {\cellcolor[HTML]{D2EDCC}} \color[HTML]{000000} 85.33 \\
         & \multirow[c]{2}{*}{16} & ADD & 84.29 & {\cellcolor[HTML]{D1EDCA}} \color[HTML]{000000} 76.26 & {\cellcolor[HTML]{D4EECE}} \color[HTML]{000000} 78.30 & {\cellcolor[HTML]{EDF8E7}} \color[HTML]{000000} 94.26 & {\cellcolor[HTML]{ABDEB6}} \color[HTML]{000000} 81.20 & {\cellcolor[HTML]{E1F3DC}} \color[HTML]{000000} 91.46 \\
         &  & ADD-S & 97.08 & {\cellcolor[HTML]{F4FBED}} \color[HTML]{000000} 95.65 & {\cellcolor[HTML]{F1F9EA}} \color[HTML]{000000} 94.13 & {\cellcolor[HTML]{F4FBED}} \color[HTML]{000000} 98.12 & {\cellcolor[HTML]{F7FCF0}} \color[HTML]{000000} 100.00 & {\cellcolor[HTML]{F1F9EA}} \color[HTML]{000000} 97.52 \\
         \midrule
        \multirow[c]{6}{*}{SAM6D~\cite{sam6d}} & \multirow[c]{2}{*}{1} & ADD & 29.16 & {\cellcolor[HTML]{085093}} \color[HTML]{F1F1F1} 6.13 & {\cellcolor[HTML]{1171B1}} \color[HTML]{F1F1F1} 16.49 & {\cellcolor[HTML]{3DA0C9}} \color[HTML]{F1F1F1} 33.23 & {\cellcolor[HTML]{084B8D}} \color[HTML]{F1F1F1} 50.47 & {\cellcolor[HTML]{0867AB}} \color[HTML]{F1F1F1} 39.50 \\
         &  & ADD-S & 61.69 & {\cellcolor[HTML]{0B6CAE}} \color[HTML]{F1F1F1} 14.55 & {\cellcolor[HTML]{A6DCB6}} \color[HTML]{000000} 60.82 & {\cellcolor[HTML]{ECF8E6}} \color[HTML]{000000} 94.10 & {\cellcolor[HTML]{4AAFD1}} \color[HTML]{F1F1F1} 67.20 & {\cellcolor[HTML]{9CD9B9}} \color[HTML]{000000} 71.81 \\
         & \multirow[c]{2}{*}{8} & ADD & 37.00 & {\cellcolor[HTML]{0862A5}} \color[HTML]{F1F1F1} 11.35 & {\cellcolor[HTML]{3EA1C9}} \color[HTML]{F1F1F1} 31.96 & {\cellcolor[HTML]{5ABACF}} \color[HTML]{000000} 42.13 & {\cellcolor[HTML]{085598}} \color[HTML]{F1F1F1} 52.12 & {\cellcolor[HTML]{2889BC}} \color[HTML]{F1F1F1} 47.47 \\
         &  & ADD-S & 70.35 & {\cellcolor[HTML]{43A7CD}} \color[HTML]{F1F1F1} 33.67 & {\cellcolor[HTML]{B3E1BA}} \color[HTML]{000000} 64.95 & {\cellcolor[HTML]{EEF9E8}} \color[HTML]{000000} 95.10 & {\cellcolor[HTML]{88D1C0}} \color[HTML]{000000} 76.25 & {\cellcolor[HTML]{C8EAC3}} \color[HTML]{000000} 81.81 \\
         & \multirow[c]{2}{*}{16} & ADD & 82.65 & {\cellcolor[HTML]{C0E6C0}} \color[HTML]{000000} 69.21 & {\cellcolor[HTML]{BFE6BF}} \color[HTML]{000000} 69.16 & {\cellcolor[HTML]{E1F4DC}} \color[HTML]{000000} 88.36 & {\cellcolor[HTML]{E9F7E3}} \color[HTML]{000000} 96.12 & {\cellcolor[HTML]{DEF2D9}} \color[HTML]{000000} 90.40 \\
         &  & ADD-S & 93.00 & {\cellcolor[HTML]{EAF7E4}} \color[HTML]{000000} 90.22 & {\cellcolor[HTML]{D1EDCA}} \color[HTML]{000000} 76.29 & {\cellcolor[HTML]{F7FCF0}} \color[HTML]{000000} 99.93 & {\cellcolor[HTML]{F6FBEF}} \color[HTML]{000000} 99.59 & {\cellcolor[HTML]{F5FBEE}} \color[HTML]{000000} 98.99 \\
        \midrule
        \multirow[c]{6}{*}{Ours} & \multirow[c]{2}{*}{1} & ADD & 50.11 & {\cellcolor[HTML]{5ABACF}} \color[HTML]{000000} 40.30 & {\cellcolor[HTML]{45A8CD}} \color[HTML]{F1F1F1} 34.22 & {\cellcolor[HTML]{51B5D2}} \color[HTML]{F1F1F1} 40.10 & {\cellcolor[HTML]{B9E3BC}} \color[HTML]{000000} 83.63 & {\cellcolor[HTML]{3C9FC8}} \color[HTML]{F1F1F1} 52.31 \\
         &  & ADD-S & 84.20 & {\cellcolor[HTML]{B9E3BC}} \color[HTML]{000000} 66.67 & {\cellcolor[HTML]{DFF3DA}} \color[HTML]{000000} 85.23 & {\cellcolor[HTML]{E1F4DC}} \color[HTML]{000000} 88.30 & {\cellcolor[HTML]{DCF1D6}} \color[HTML]{000000} 92.10 & {\cellcolor[HTML]{DAF1D5}} \color[HTML]{000000} 88.70 \\
         & \multirow[c]{2}{*}{8} & ADD & 67.27 & {\cellcolor[HTML]{86D0C0}} \color[HTML]{000000} 52.21 & {\cellcolor[HTML]{86D0C0}} \color[HTML]{000000} 52.10 & {\cellcolor[HTML]{BFE6BF}} \color[HTML]{000000} 71.10 & {\cellcolor[HTML]{C9EAC4}} \color[HTML]{000000} 86.67 & {\cellcolor[HTML]{A9DDB5}} \color[HTML]{000000} 74.27 \\
         &  & ADD-S & 91.58 & {\cellcolor[HTML]{E5F5E0}} \color[HTML]{000000} 87.70 & {\cellcolor[HTML]{E9F6E3}} \color[HTML]{000000} 90.12 & {\cellcolor[HTML]{E5F5E0}} \color[HTML]{000000} 90.20 & {\cellcolor[HTML]{EEF8E7}} \color[HTML]{000000} 97.30 & {\cellcolor[HTML]{E4F4DE}} \color[HTML]{000000} 92.62 \\
         & \multirow[c]{2}{*}{16} & ADD & 85.00 & {\cellcolor[HTML]{D3EECC}} \color[HTML]{000000} 77.21 & {\cellcolor[HTML]{D8F0D3}} \color[HTML]{000000} 81.10 & {\cellcolor[HTML]{E1F4DC}} \color[HTML]{000000} 88.20 & {\cellcolor[HTML]{D6EFD0}} \color[HTML]{000000} 90.21 & {\cellcolor[HTML]{D9F0D3}} \color[HTML]{000000} 88.30 \\
         &  & ADD-S & 98.19 & {\cellcolor[HTML]{F7FCF0}} \color[HTML]{000000} 97.34 & {\cellcolor[HTML]{F3FAEC}} \color[HTML]{000000} 95.33 & {\cellcolor[HTML]{F4FBED}} \color[HTML]{000000} 98.30 & {\cellcolor[HTML]{F7FCF0}} \color[HTML]{000000} 100.00 & {\cellcolor[HTML]{F7FCF0}} \color[HTML]{000000} 100.00 \\
        \bottomrule
        \end{tabular}
    }
    \caption{Quantitative performance of~\method compared to SOTA baselines on LM-O~\cite{lmo} for 6D Pose estimation. For both ADD and ADD-S metrics, 
    we show higher values with brighter gradient and vice-versa.}
    \label{tab:lmo_pose}
\end{table}

\begin{figure}[h!]
    \centering
    \includegraphics[width=\textwidth]{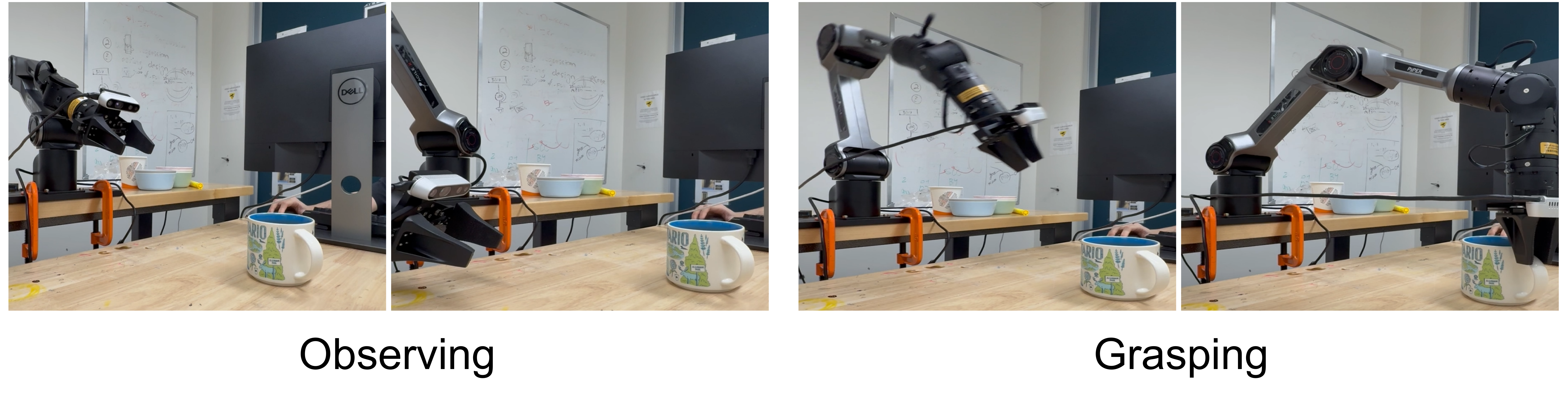}
    \caption{Visualization of~\method for a real-world robotic manipulation task where it estimates the complete geometry and 6D pose of the cup.}
    \label{fig:robot_demo}
\end{figure}

\section{Deployments on a Real-world Robotic Platform}
\label{app:real_world_robot}

We further demonstrate the effectiveness and the deployment capablity of~\method in a real-world robotic manipulation pipeline using a PiPER robot arm. In this setup, \method estimates the complete 3D goemetry and 6D pose of a target object from RGB-D stream inputs captured by a wrist-mounted Intel RealSense D435 camera. This geometric and pose information is then passed to a subsequent grasp planning module, AnyGrasp~\cite{fang2023anygrasp}, to determine a suitable grasp pose for the robot. 
\cref{fig:robot_demo} provides a visual snapshot of this system in action, showcasing \method's reconstruction and pose estimation for a target mug on PiPER. 
Further details, including the dynamic operation of the system capturing multiple views, the resulting reconstruction, and the robot successfully grasping the object, are presented in the accompanying video.

\begin{table}[h!]
\centering
\resizebox{\textwidth}{!}{
\scriptsize
\begin{tabular}{c|c|c|l|*{6}{c}}
\toprule
Method & \# img & Metric &  Mean & $\text{005\_tomato\_soup\_can}$ & $\text{006\_mustard\_bottle}$ & $\text{007\_tuna\_fish\_can}$ & $\text{009\_gelatin\_box}$ & $\text{010\_potted\_meat\_can}$ & $\text{011\_banana}$ \\
\midrule

 \multirow[c]{6}{*}{GigaPose~\cite{gigapose}} & \multirow[c]{2}{*}{1} & ADD & 13.58 & {\cellcolor[HTML]{085598}} \color[HTML]{F1F1F1} 8.20 & {\cellcolor[HTML]{084081}} \color[HTML]{F1F1F1} 4.15 & {\cellcolor[HTML]{084081}} \color[HTML]{F1F1F1} 1.77 & {\cellcolor[HTML]{1F80B8}} \color[HTML]{F1F1F1} 45.92 & {\cellcolor[HTML]{084081}} \color[HTML]{F1F1F1} 4.72 & {\cellcolor[HTML]{085093}} \color[HTML]{F1F1F1} 16.74 \\
 &  & ADD-S & 24.44 & {\cellcolor[HTML]{2B8CBE}} \color[HTML]{F1F1F1} 26.23 & {\cellcolor[HTML]{085598}} \color[HTML]{F1F1F1} 10.66 & {\cellcolor[HTML]{084B8D}} \color[HTML]{F1F1F1} 5.42 & {\cellcolor[HTML]{80CEC2}} \color[HTML]{000000} 66.69 & {\cellcolor[HTML]{085FA3}} \color[HTML]{F1F1F1} 14.17 & {\cellcolor[HTML]{0868AC}} \color[HTML]{F1F1F1} 23.48 \\
 & \multirow[c]{2}{*}{8} & ADD & 21.48 & {\cellcolor[HTML]{1373B2}} \color[HTML]{F1F1F1} 17.74 & {\cellcolor[HTML]{0861A4}} \color[HTML]{F1F1F1} 14.03 & {\cellcolor[HTML]{084586}} \color[HTML]{F1F1F1} 3.33 & {\cellcolor[HTML]{ABDEB6}} \color[HTML]{000000} 75.00 & {\cellcolor[HTML]{084688}} \color[HTML]{F1F1F1} 6.62 & {\cellcolor[HTML]{084081}} \color[HTML]{F1F1F1} 12.16 \\
 &  & ADD-S & 39.52 & {\cellcolor[HTML]{99D7BA}} \color[HTML]{000000} 58.06 & {\cellcolor[HTML]{2D8FBF}} \color[HTML]{F1F1F1} 28.87 & {\cellcolor[HTML]{0868AC}} \color[HTML]{F1F1F1} 14.35 & {\cellcolor[HTML]{BEE6BF}} \color[HTML]{000000} 79.62 & {\cellcolor[HTML]{1979B5}} \color[HTML]{F1F1F1} 22.47 & {\cellcolor[HTML]{2A8BBE}} \color[HTML]{F1F1F1} 33.78 \\
 & \multirow[c]{2}{*}{16} & ADD & 60.58 & {\cellcolor[HTML]{E5F5E0}} \color[HTML]{000000} 88.80 & {\cellcolor[HTML]{A2DBB7}} \color[HTML]{000000} 62.38 & {\cellcolor[HTML]{0867AB}} \color[HTML]{F1F1F1} 13.67 & {\cellcolor[HTML]{E6F6E1}} \color[HTML]{000000} 93.75 & {\cellcolor[HTML]{A0DAB8}} \color[HTML]{000000} 62.35 & {\cellcolor[HTML]{46AACE}} \color[HTML]{F1F1F1} 42.56 \\
 &  & ADD-S & 86.8 & {\cellcolor[HTML]{F4FBED}} \color[HTML]{000000} 96.82 & {\cellcolor[HTML]{F7FCF0}} \color[HTML]{000000} 100.00 & {\cellcolor[HTML]{75C8C6}} \color[HTML]{000000} 49.33 & {\cellcolor[HTML]{F7FCF0}} \color[HTML]{000000} 100.00 & {\cellcolor[HTML]{E9F6E3}} \color[HTML]{000000} 92.35 & {\cellcolor[HTML]{D4EECE}} \color[HTML]{000000} 82.33 \\

 \midrule
\multirow[c]{6}{*}{FoundationPose~\cite{fp}} & \multirow[c]{2}{*}{1} & ADD & 30.34 & {\cellcolor[HTML]{084081}} \color[HTML]{F1F1F1} 1.69 & {\cellcolor[HTML]{4DB2D3}} \color[HTML]{F1F1F1} 40.00 & {\cellcolor[HTML]{085395}} \color[HTML]{F1F1F1} 7.69 & {\cellcolor[HTML]{AFE0B8}} \color[HTML]{000000} 76.00 & {\cellcolor[HTML]{0868AC}} \color[HTML]{F1F1F1} 16.67 & {\cellcolor[HTML]{3EA1C9}} \color[HTML]{F1F1F1} 40.00 \\
 &  & ADD-S & 74.58 & {\cellcolor[HTML]{5DBBCE}} \color[HTML]{000000} 42.01 & {\cellcolor[HTML]{DFF2DA}} \color[HTML]{000000} 86.90 & {\cellcolor[HTML]{BFE6BF}} \color[HTML]{000000} 71.10 & {\cellcolor[HTML]{F7FCF0}} \color[HTML]{000000} 100.00 & {\cellcolor[HTML]{C0E6C0}} \color[HTML]{000000} 72.10 & {\cellcolor[HTML]{C4E8C1}} \color[HTML]{000000} 75.40 \\
 & \multirow[c]{2}{*}{8} & ADD & 45.57 & {\cellcolor[HTML]{65C0CB}} \color[HTML]{000000} 44.07 & {\cellcolor[HTML]{86D0C0}} \color[HTML]{000000} 55.00 & {\cellcolor[HTML]{57B8D0}} \color[HTML]{000000} 41.00 & {\cellcolor[HTML]{C0E6C0}} \color[HTML]{000000} 80.00 & {\cellcolor[HTML]{085DA0}} \color[HTML]{F1F1F1} 13.33 & {\cellcolor[HTML]{3EA1C9}} \color[HTML]{F1F1F1} 40.00 \\
 &  & ADD-S & 90.17 & {\cellcolor[HTML]{D0ECC9}} \color[HTML]{000000} 76.20 & {\cellcolor[HTML]{DFF3DA}} \color[HTML]{000000} 87.33 & {\cellcolor[HTML]{E7F6E2}} \color[HTML]{000000} 91.20 & {\cellcolor[HTML]{F7FCF0}} \color[HTML]{000000} 100.00 & {\cellcolor[HTML]{E2F4DD}} \color[HTML]{000000} 89.10 & {\cellcolor[HTML]{F1FAEB}} \color[HTML]{000000} 97.20 \\
 & \multirow[c]{2}{*}{16} & ADD & 71.2 & {\cellcolor[HTML]{A7DDB5}} \color[HTML]{000000} 62.10 & {\cellcolor[HTML]{AEDFB8}} \color[HTML]{000000} 66.00 & {\cellcolor[HTML]{C3E7C1}} \color[HTML]{000000} 72.30 & {\cellcolor[HTML]{F7FCF0}} \color[HTML]{000000} 100.00 & {\cellcolor[HTML]{9BD8B9}} \color[HTML]{000000} 60.60 & {\cellcolor[HTML]{A5DCB6}} \color[HTML]{000000} 66.20 \\
 &  & ADD-S & 95.56 & {\cellcolor[HTML]{F7FCF0}} \color[HTML]{000000} 98.37 & {\cellcolor[HTML]{EEF8E7}} \color[HTML]{000000} 95.00 & {\cellcolor[HTML]{F7FCF0}} \color[HTML]{000000} 100.00 & {\cellcolor[HTML]{F7FCF0}} \color[HTML]{000000} 100.00 & {\cellcolor[HTML]{D3EECC}} \color[HTML]{000000} 80.00 & {\cellcolor[HTML]{F7FCF0}} \color[HTML]{000000} 100.00 \\
 \midrule
\multirow[c]{6}{*}{SAM6D~\cite{sam6d}} & \multirow[c]{2}{*}{1} & ADD & 19.93 & {\cellcolor[HTML]{0861A4}} \color[HTML]{F1F1F1} 11.66 & {\cellcolor[HTML]{47ABCF}} \color[HTML]{F1F1F1} 37.50 & {\cellcolor[HTML]{084688}} \color[HTML]{F1F1F1} 3.75 & {\cellcolor[HTML]{084081}} \color[HTML]{F1F1F1} 31.60 & {\cellcolor[HTML]{085294}} \color[HTML]{F1F1F1} 10.00 & {\cellcolor[HTML]{0E6EAF}} \color[HTML]{F1F1F1} 25.10 \\
 &  & ADD-S & 43.04 & {\cellcolor[HTML]{0866A9}} \color[HTML]{F1F1F1} 13.33 & {\cellcolor[HTML]{DEF2D9}} \color[HTML]{000000} 86.83 & {\cellcolor[HTML]{1070B0}} \color[HTML]{F1F1F1} 16.76 & {\cellcolor[HTML]{389BC6}} \color[HTML]{F1F1F1} 52.10 & {\cellcolor[HTML]{72C7C7}} \color[HTML]{000000} 50.00 & {\cellcolor[HTML]{3B9DC7}} \color[HTML]{F1F1F1} 39.20 \\
 & \multirow[c]{2}{*}{8} & ADD & 34.63 & {\cellcolor[HTML]{1070B0}} \color[HTML]{F1F1F1} 16.67 & {\cellcolor[HTML]{5FBDCD}} \color[HTML]{000000} 44.67 & {\cellcolor[HTML]{085093}} \color[HTML]{F1F1F1} 6.87 & {\cellcolor[HTML]{B1E0B9}} \color[HTML]{000000} 76.25 & {\cellcolor[HTML]{48ACCF}} \color[HTML]{F1F1F1} 38.33 & {\cellcolor[HTML]{0E6EAF}} \color[HTML]{F1F1F1} 25.00 \\
 &  & ADD-S & 63.86 & {\cellcolor[HTML]{4AAFD1}} \color[HTML]{F1F1F1} 36.67 & {\cellcolor[HTML]{D7EFD1}} \color[HTML]{000000} 82.33 & {\cellcolor[HTML]{389BC6}} \color[HTML]{F1F1F1} 31.00 & {\cellcolor[HTML]{EEF9E8}} \color[HTML]{000000} 96.70 & {\cellcolor[HTML]{ABDEB6}} \color[HTML]{000000} 65.17 & {\cellcolor[HTML]{B6E3BB}} \color[HTML]{000000} 71.28 \\
 & \multirow[c]{2}{*}{16} & ADD & 72.06 & {\cellcolor[HTML]{A2DBB7}} \color[HTML]{000000} 60.27 & {\cellcolor[HTML]{D3EECD}} \color[HTML]{000000} 80.20 & {\cellcolor[HTML]{CFECC8}} \color[HTML]{000000} 77.10 & {\cellcolor[HTML]{F7FCF0}} \color[HTML]{000000} 100.00 & {\cellcolor[HTML]{8AD2BF}} \color[HTML]{000000} 56.36 & {\cellcolor[HTML]{84CFC1}} \color[HTML]{000000} 58.44 \\
 &  & ADD-S & 97.02 & {\cellcolor[HTML]{F7FCF0}} \color[HTML]{000000} 98.42 & {\cellcolor[HTML]{F7FCF0}} \color[HTML]{000000} 100.00 & {\cellcolor[HTML]{F7FCF0}} \color[HTML]{000000} 100.00 & {\cellcolor[HTML]{F7FCF0}} \color[HTML]{000000} 100.00 & {\cellcolor[HTML]{D9F0D3}} \color[HTML]{000000} 83.73 & {\cellcolor[HTML]{F7FCF0}} \color[HTML]{000000} 100.00 \\
 \midrule
\multirow[c]{6}{*}{Ours} & \multirow[c]{2}{*}{1} & ADD & 47.92 & {\cellcolor[HTML]{1B7BB6}} \color[HTML]{F1F1F1} 20.34 & {\cellcolor[HTML]{86D0C0}} \color[HTML]{000000} 55.00 & {\cellcolor[HTML]{4CB1D2}} \color[HTML]{F1F1F1} 38.20 & {\cellcolor[HTML]{C0E6C0}} \color[HTML]{000000} 80.00 & {\cellcolor[HTML]{5BBACF}} \color[HTML]{000000} 44.00 & {\cellcolor[HTML]{62BECC}} \color[HTML]{000000} 50.00 \\
 &  & ADD-S & 82.72 & {\cellcolor[HTML]{D9F0D3}} \color[HTML]{000000} 82.12 & {\cellcolor[HTML]{E4F5DF}} \color[HTML]{000000} 90.00 & {\cellcolor[HTML]{BAE4BD}} \color[HTML]{000000} 69.23 & {\cellcolor[HTML]{F7FCF0}} \color[HTML]{000000} 100.00 & {\cellcolor[HTML]{D3EECC}} \color[HTML]{000000} 80.00 & {\cellcolor[HTML]{C3E7C1}} \color[HTML]{000000} 75.00 \\
 & \multirow[c]{2}{*}{8} & ADD & 61.2 & {\cellcolor[HTML]{94D5BC}} \color[HTML]{000000} 56.60 & {\cellcolor[HTML]{BAE4BD}} \color[HTML]{000000} 70.00 & {\cellcolor[HTML]{86D0C0}} \color[HTML]{000000} 53.60 & {\cellcolor[HTML]{D6EFD0}} \color[HTML]{000000} 87.00 & {\cellcolor[HTML]{4CB1D2}} \color[HTML]{F1F1F1} 40.00 & {\cellcolor[HTML]{8BD2BF}} \color[HTML]{000000} 60.00 \\
 &  & ADD-S & 91.22 & {\cellcolor[HTML]{E4F5DF}} \color[HTML]{000000} 88.22 & {\cellcolor[HTML]{F7FCF0}} \color[HTML]{000000} 100.00 & {\cellcolor[HTML]{E3F4DE}} \color[HTML]{000000} 89.13 & {\cellcolor[HTML]{F7FCF0}} \color[HTML]{000000} 100.00 & {\cellcolor[HTML]{E4F5DF}} \color[HTML]{000000} 90.00 & {\cellcolor[HTML]{D0ECC9}} \color[HTML]{000000} 80.00 \\
 & \multirow[c]{2}{*}{16} & ADD & 74.39 & {\cellcolor[HTML]{B1E0B9}} \color[HTML]{000000} 64.98 & {\cellcolor[HTML]{D3EECC}} \color[HTML]{000000} 80.00 & {\cellcolor[HTML]{B3E1BA}} \color[HTML]{000000} 66.67 & {\cellcolor[HTML]{DFF3DA}} \color[HTML]{000000} 91.00 & {\cellcolor[HTML]{CDEBC6}} \color[HTML]{000000} 76.67 & {\cellcolor[HTML]{A7DDB5}} \color[HTML]{000000} 67.00 \\
 &  & ADD-S & 96.65 & {\cellcolor[HTML]{E7F6E2}} \color[HTML]{000000} 89.83 & {\cellcolor[HTML]{F7FCF0}} \color[HTML]{000000} 100.00 & {\cellcolor[HTML]{EEF8E7}} \color[HTML]{000000} 94.87 & {\cellcolor[HTML]{F7FCF0}} \color[HTML]{000000} 100.00 & {\cellcolor[HTML]{F7FCF0}} \color[HTML]{000000} 100.00 & {\cellcolor[HTML]{EEF8E7}} \color[HTML]{000000} 95.20 \\ 
 
\bottomrule
\end{tabular}
}
\caption{Quantitative performance of~\method compared to SOTA baselines on YCB-Video~\cite{ycbv} for 6D Pose estimation. For both ADD and ADD-S metrics, 
we show higher values with brighter gradient and vice-versa.}
\label{tab:ycb_pose}
\end{table}

\end{document}

%% file: notation.tex
\usepackage{xifthen}
\usepackage{xparse}
\usepackage{subdepth}
\usepackage{leftidx}
\usepackage{bm}
\usepackage{amssymb,amsmath}
\usepackage{amsfonts}

\newcommand{\bbm}{\begin{bmatrix}}
	\newcommand{\ebm}{\end{bmatrix}}

\DeclareMathAlphabet{\mbf}{OT1}{ptm}{b}{n}
\newcommand{\mbs}[1]{{\bm{#1}}}

\newcommand{\mbsbar}[1]{{\overline{\boldsymbol{#1}}}}
\newcommand{\mbshat}[1]{{\hat{\boldsymbol{#1}}}}
\newcommand{\mbstilde}[1]{{\tilde{\boldsymbol{#1}}}}

\newcommand{\mbsdot}[1]{{\dot {\boldsymbol{#1}}}}

\newcommand{\mbc}[1]{{{\boldsymbol{\mathcal{#1}}}}}

\newcommand{\mbfbar}[1]{{\overline{\mbf{#1}}}}
\newcommand{\mbfhat}[1]{{\hat{\mbf{#1}}}}
\newcommand{\mbftilde}[1]{{\tilde{\mbf{#1}}}}

\newcommand{\mbfdot}[1]{{\dot{\mbf{#1}}}}

\newcommand{\cframe}[1]{{\smash{\protect\underrightarrow{\mathcal{F}}_{#1}}}}

\DeclareMathAlphabet{\mathbfit}{OML}{cmm}{b}{it}
\newcommand{\homo}[1]{{\mathbfit{#1}}}

\newcommand{\mbfh}[1]{{\homo{#1}}}

\newcommand{\trans}[3]{\leftidx{_{#1}}{\mbf r}{\IfValueTF{#2}{_{#2#3\hspace{2pt}}}{}}} %

\newcommand{\vel}[3]{\leftidx{_{#1}}{\mbf v}{\IfValueTF{#2}{_{#2#3\hspace{2pt}}}{}}} %
\newcommand{\veltilde}[3]{\leftidx{_{#1}}{\mbftilde v}{\IfValueTF{#2}{_{#2#3\hspace{2pt}}}{}}} %
\newcommand{\velbar}[3]{\leftidx{_{#1}}{\mbfbar v}{\IfValueTF{#2}{_{#2#3\hspace{2pt}}}{}}} %
\newcommand{\velhat}[3]{\leftidx{_{#1}}{\mbfhat v}{\IfValueTF{#2}{_{#2#3\hspace{2pt}}}{}}} %
\newcommand{\veldot}[3]{\leftidx{_{#1}}{\mbfdot v}{\IfValueTF{#2}{_{#2#3\hspace{2pt}}}{}}} %

\newcommand{\acc}[3]{\leftidx{_{#1}}{\mbf a}{\IfValueTF{#2}{_{#2#3\hspace{2pt}}}{}}} %
\newcommand{\acctilde}[3]{\leftidx{_{#1}}{\mbftilde a}{\IfValueTF{#2}{_{#2#3\hspace{2pt}}}{}}} %
\newcommand{\accbar}[3]{\leftidx{_{#1}}{\mbfbar a}{\IfValueTF{#2}{_{#2#3\hspace{2pt}}}{}}} %

\newcommand{\rotvel}[3]{\leftidx{_{#1}}{\mbs \omega}{\IfValueTF{#2}{_{#2#3\hspace{2pt}}}{}}} %
\newcommand{\rotveltilde}[3]{\leftidx{_{#1}}{\mbstilde \omega}{\IfValueTF{#2}{_{#2#3\hspace{2pt}}}{}}} %
\newcommand{\rotvelbar}[3]{\leftidx{_{#1}}{\mbsbar \omega}{\IfValueTF{#2}{_{#2#3\hspace{2pt}}}{}}} %
\newcommand{\rotvelhat}[3]{\leftidx{_{#1}}{\mbshat \omega}{\IfValueTF{#2}{_{#2#3\hspace{2pt}}}{}}} %
\newcommand{\rotveldot}[3]{\leftidx{_{#1}}{\mbsdot \omega}{\IfValueTF{#2}{_{#2#3\hspace{2pt}}}{}}} %

\newcommand{\T}[2]{\leftidx{}{\mbfh T}{_{#1#2\hspace{2pt}}}} %

%% file: intro.tex
\label{sec:intro}

6D object pose estimation is a fundamental challenge in robotic manipulation, enabling tasks from precision grasping to complex assembly~\cite{zhou2024deep,sundermeyer2021contact}.
While deep learning (DL) has driven notable progress, 
prior approaches have predominantly relied on object-specific textured CAD models~\cite{park2019pix2pose,labbe2020cosypose,he2020pvn3d,he2021ffb6d} or category-level training data~\cite{wang2019normalized,zhang2024omni6dpose,chen2020learning,lin2024instance,chen2024secondpose}, severely limiting their application in dynamic, open-world environments. 
Even recent category-agnostic approaches~\cite{caraffa2024freeze,sam6d,chen2024zeropose} still depend on ground-truth CAD models during inference, essentially approximating the model-based paradigm rather than transcending it. 
As robots increasingly operate in open-world environments with novel objects, the need for methods that can estimate poses without prior object models has become increasingly urgent.
Model-free pose estimation methods offer a promising alternative by eliminating dependence on pre-existing object models, instead leveraging reference views~\cite{sam6d,onepose,sun2021loftr,nguyen2024nope}, or image-to-3D reconstruction models~\cite{liu2022gen6d,lee2025any6d,gigapose,liu2025hippoharnessingimageto3dpriors}.
However, despite their ``model-free'' designation, these methods still implicitly rely on reference object models generated in advance from either diffusion models or multiple reference views of the target objects.
Therefore, a key issue arises when 3D representations are generated from limited observations: unobserved regions introduce variable confidence~\ie, epistemic uncertainty during reconstruction.
Existing approaches typically fail to quantify this epistemic uncertainty~\cite{kendall2017uncertainties}, treating all generated regions equally regardless of observability.
This limitation becomes particularly problematic when integrating new observations, as the system cannot appropriately weigh the reliability of the prior model against newly observed data~\cite{Thrun:etal:Book2005,gom}.

To this end,
we introduce \method, a principled framework for zero-shot, model-free 6D object pose estimation that addresses these limitations through uncertainty-aware multi-view integration and pose graph optimization.
Starting from a single RGB-D observation, \method generates an initial 3D representation using an image-to-3D diffusion model with pixel-level uncertainty estimates that quantify confidence in the diffusion priors and balance them with sensor measurements.
This 3D representation is encoded using 3D Gaussian Splatting (3DGS)~\cite{3dgs}, enabling efficient and high-fidelity textured models. 
As more observations arrive, \method incrementally refines the geometry and appearance via pose graph optimization, with uncertainty estimates guiding the fusion process to prioritize reliable measurements while refining uncertain regions.

In summary, our contributions are: 
(1) we present a model-free approach that eliminates dependencies on object-specific CAD models, category-level training, or multi-view requirements with known camera poses;
(2) we introduce an uncertainty-guided refinement that adaptively integrates observed data into diffusion priors; 
(3) We formulate pose estimation as an incremental factor graph optimization problem that incorporates both diffusion priors and observations, and ensures global consistency; and 
(4) We demonstrate significant improvements in both pose estimation accuracy and reconstruction quality compared to existing SOTA methods.